%% file: auto_exposure.tex
\documentclass[10pt,journal,compsoc]{IEEEtran}

\ifCLASSOPTIONcompsoc
  \usepackage[nocompress]{cite}
\else
  \usepackage{cite}
\fi
\ifCLASSINFOpdf
\else
\fi
\usepackage{booktabs} 
\usepackage{amsmath,amsfonts,amssymb}
\usepackage{lipsum}
\usepackage{color}
\usepackage{url}
\usepackage{algorithmic}
\usepackage{graphicx}
\usepackage{caption}
\usepackage{subcaption}
\usepackage{verbatim}
\usepackage{float}
\usepackage{multirow}
\usepackage{rotating}
\usepackage[normalem]{ulem}

\newcommand{\changed}[1]{\textcolor{black}{#1}}

\usepackage[ruled]{algorithm2e} 

\begin{document}
%
\title{Personalized Exposure Control Using \changed{Adaptive Metering and} Reinforcement Learning}
%
%
%
%

\author{Huan Yang,
        Baoyuan Wang$^*$\thanks{* Corresponding Author},
        Noranart Vesdapunt, \\
        Minyi Guo~\IEEEmembership{Fellow,~IEEE} and
        Sing Bing Kang~\IEEEmembership{Fellow,~IEEE}
\IEEEcompsocitemizethanks{\IEEEcompsocthanksitem Huan Yang and Minyi Guo is with the Department
of Computer Science and Engineering, Shanghai Jiao Tong University, China.\protect\\
\IEEEcompsocthanksitem Baoyuan Wang, Noranart Vesdapunt and Sing Bing Kang are with Microsoft Research.}

}

%
%

\markboth{Journal of \LaTeX\ Class Files,~Vol.~14, No.~8, August~2015}%
{Shell \MakeLowercase{\textit{et al.}}: Bare Advanced Demo of IEEEtran.cls for IEEE Computer Society Journals}
%



\IEEEtitleabstractindextext{%
\input{abstract}

\begin{IEEEkeywords}
Auto Exposure, Reinforcement Learning, Personalization
\end{IEEEkeywords}}

\maketitle

\IEEEdisplaynontitleabstractindextext

%
\IEEEpeerreviewmaketitle

\input{introduction_related}


\input{overview}
\input{attention}

\input{reinforce}

\input{experiment}

\input{conclusion}

\appendices
\input{appendix}

\ifCLASSOPTIONcompsoc
  \section*{Acknowledgments}
\else
  \section*{Acknowledgment}
\fi

This work was partially done when the first author was an intern at Microsoft AI\&R.

\ifCLASSOPTIONcaptionsoff
  \newpage
\fi



%




\bibliographystyle{IEEEtran}
\bibliography{AE_DNN}

%








\begin{IEEEbiography}[{\includegraphics[width=1in,height=1.25in,clip,keepaspectratio]{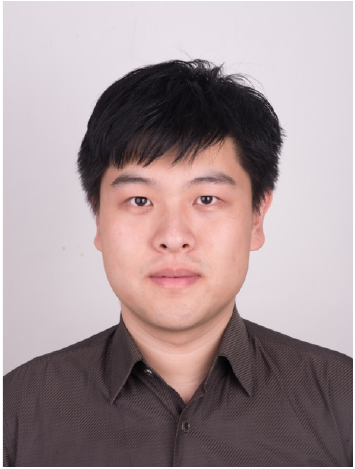}}]{Huan Yang}
received the BS degree in computer science in 2014 from Shanghai Jiao Tong University, China. He is currently pursuing the Ph.D. degree in computer science at Shanghai Jiao Tong University. His current research interests include deep learning, reinforcement learning, computer vision, image processing, real-time video processing and image photography.
\end{IEEEbiography}

\begin{IEEEbiography}
[{\includegraphics[width=1in,height=1.25in,clip,keepaspectratio]{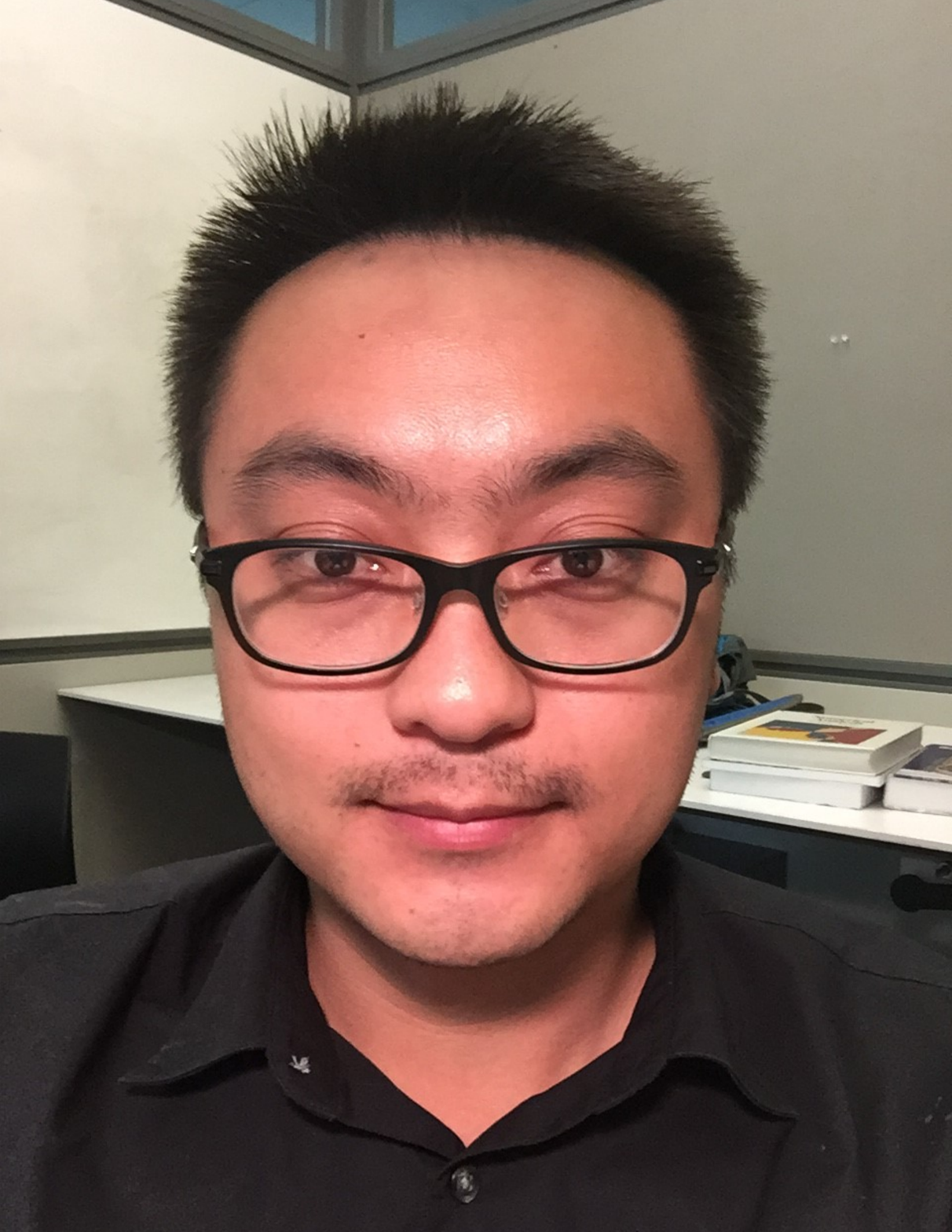}}]
{Baoyuan Wang}
received his BS and Ph.D. degree in computer science and engineer  from Zhejiang University in 2007 and 2012 respectively. He is currently a principal researcher at Microsoft AI team, and his research interests include deep learning based computational photography for intelligent capture, intelligent action as well as  novel content generation. He has shipped several important technologies to Microsoft Pix, Xbox One, Swift-Key and other Microsoft products. From 2006 to 2007, he was an engineer intern at Infosys, Bangalore, India. From 2009 to 2012, he was an research intern at Microsoft Research Asia, working on data-driven methods for computational photography. He was a lead researcher at Microsoft Research Asia from 2012 to 2015.
\end{IEEEbiography}
\begin{IEEEbiography}[{\includegraphics[width=1in,height=1.25in,clip,keepaspectratio]{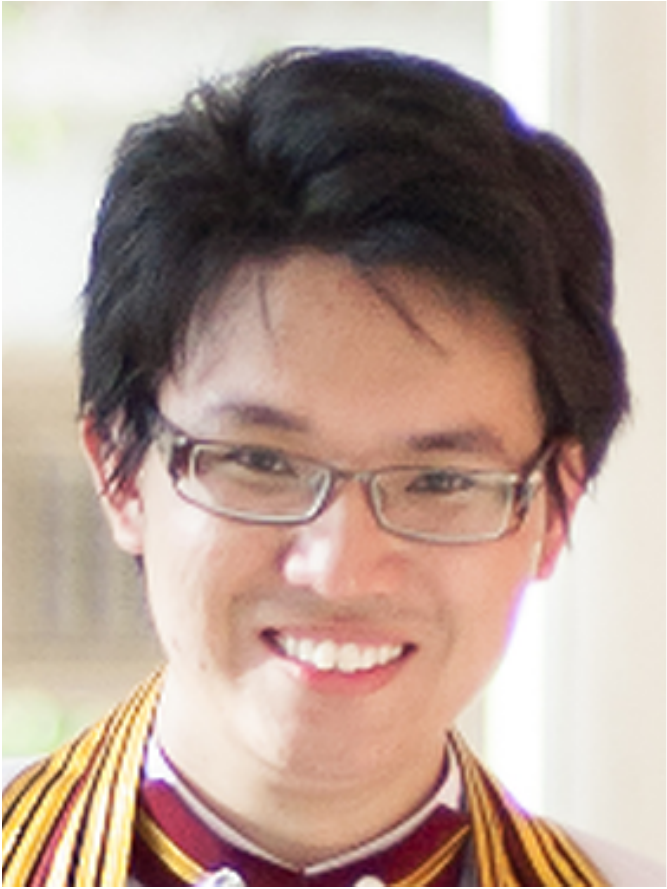}}]{Noranart Vesdapunt} received his BEng in Computer Engineering from Chulalongkorn University and MS in Computer Vision from Carnegie Mellon University. He is a software engineer at Microsoft Research. His main areas of research are computational photography, visual recognition, video analytics, and deep learning on mobile devices.
\end{IEEEbiography}
\begin{IEEEbiography}[{\includegraphics[width=1in,height=1.25in,clip,keepaspectratio]{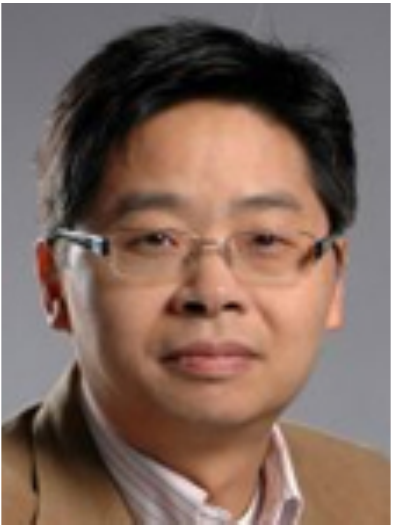}}]{Minyi Guo}
received the BS and ME degrees in computer science from Nanjing University, China, in 1982 and 1986, respectively, and the PhD degree in information science from the University of Tsukuba, Japan, in 1998. From 1998 to 2000, he had been a research associate of NEC Soft, Ltd. Japan. He was a visiting professor at the Department of Computer Science, Georgia Institute of Technology. He was a full professor at the University of Aizu, Japan, and is the head of the Department of Computer Science and Engineering at Shanghai Jiao Tong University, China. His research interests include automatic parallelization and data-parallel languages, bioinformatics, compiler optimization, high-performance computing, deep learning, and pervasive computing. He is a fellow of the IEEE.
\end{IEEEbiography}
\begin{IEEEbiography}[{\includegraphics[width=1in,height=1.25in,clip,keepaspectratio]{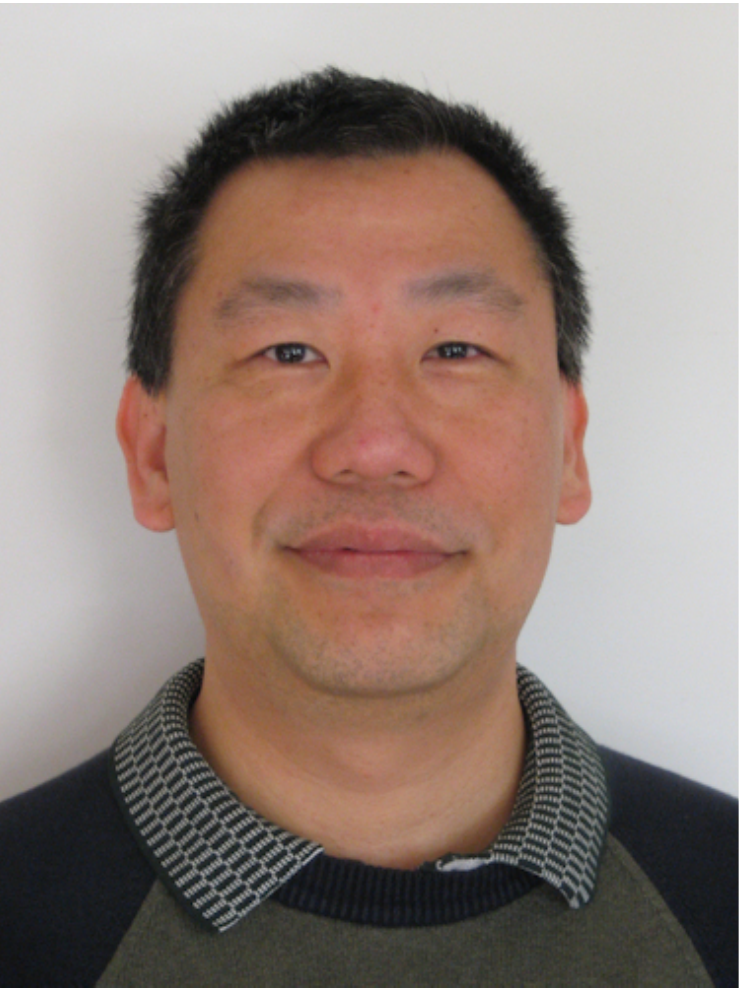}}]{Sing Bing Kang}
received his Ph.D. degree in robotics from Carnegie Mellon University, Pittsburgh, in 1994. He is Principal Researcher at Microsoft Corporation, and his research interests include image and video enhancement, and image-based modeling. He has coedited two books (Panoramic Vision and Emerging Topics in Computer Vision) and coauthored two books (Image-Based Rendering and Image-Based Modeling of Plants and Trees). On the community service front, he has served as Area Chair for the major computer vision conferences and as papers committee member for SIGGRAPH and SIGGRAPH Asia. He was Program Chair for ACCV 2007 and CVPR 2009, and was Associate Editor-In-Chief for IEEE Transactions on Pattern Analysis and Machine Intelligence from 2010-2014. He is a fellow of the IEEE.
\end{IEEEbiography}

\end{document}

%% file: abstract.tex
\begin{abstract}
We propose a reinforcement learning approach for real-time exposure control of a mobile camera that is personalizable. Our approach is based on Markov Decision Process (MDP). In the camera viewfinder or live preview mode, given the current frame, our system predicts the change in exposure so as to optimize the trade-off among image quality, fast convergence, and minimal temporal oscillation. We model the exposure prediction function as a fully convolutional neural network that can be trained through Gaussian policy gradient in an end-to-end fashion. As a result, our system can associate scene semantics with exposure values; it can also be extended to personalize the exposure adjustments for a user and device. We improve the learning performance by incorporating an \changed{adaptive metering} module that links semantics with exposure. This \changed{adaptive metering} module generalizes the conventional spot or matrix metering techniques. We validate our system using the MIT FiveK~\cite{MIT_FIVEK:2011} and our own datasets captured using iPhone 7 and Google Pixel. Experimental results show that our system exhibits stable real-time behavior while improving visual quality compared to what is achieved through native camera control.
\end{abstract}

%% file: introduction_related.tex
\ifCLASSOPTIONcompsoc
\IEEEraisesectionheading{\section{Introduction}\label{sec:introduction}}
\else
\section{Introduction}
\label{sec:introduction}
\fi

%
%
%
%
\IEEEPARstart{R}{eal-time} auto-exposure is a camera operation that enables high-quality photo capture. In smartphone cameras, this fundamental operation is typically based on simple metering over a predefined area or areas, and the analysis is independent of the scene. The newer smartphone cameras are capable of real-time detection of faces, and thus capable of using facial information to influence the exposure setting. This is especially useful for capturing backlit scenes. There are new high-end smart phones that claim to be able to detect scene categories beyond faces (e.g., Huawei P20 Pro and Xiaomi Mi 8). Details of how scene information is utilized for improving exposure control are not available, and it is not clear if the user personalization feature is available as well. 



\begin{figure}[!t]
	\centering
	\includegraphics[width=\linewidth]{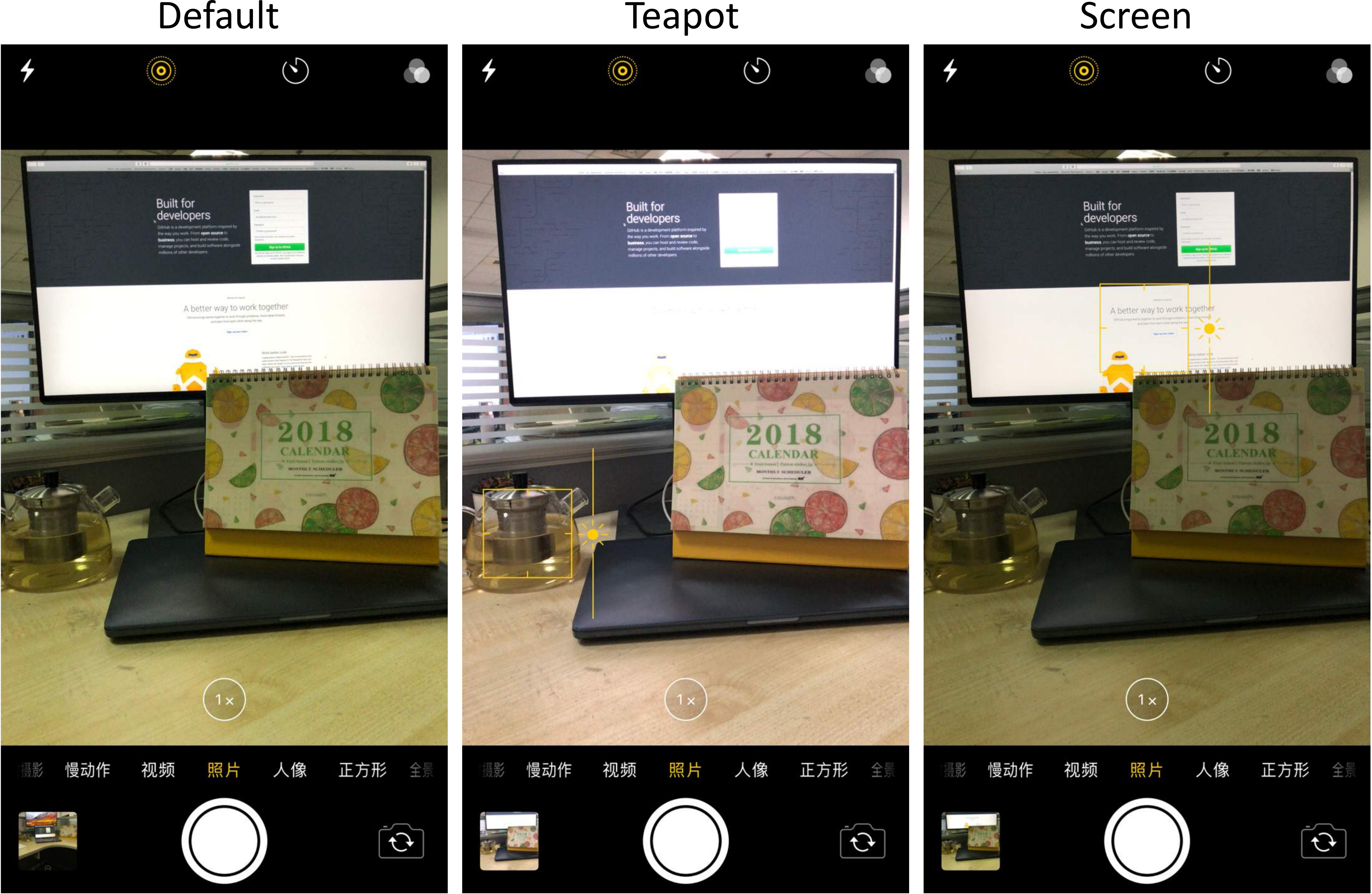}
	\caption{The ``tap-and-tweak'' strategy for manually optimizing exposure. The user selects the region of importance by tapping on it, and the camera responds by prioritizing exposure over that region.}
	\label{fig:iphone_capture}
\end{figure}

Currently, if the user is not satisfied with the viewfinder (live preview) exposure and wishes to modify it, he/she would need to tap on the region of interest and then manually tweak the exposure by the sliding bar before the shutter press, as shown in Figure~\ref{fig:iphone_capture}. However, during this ``tap-and-tweak'' process, it would be easy to miss the best moment to capture. Our goal is develop a system that automatically produces the exposure that is acceptable to the user and thus obviates the need for such manual adjustment. 

We are not aware of published research work done on using generic scene information for real-time exposure control, not to mention additionally catering to specific user preferences. It is evident from the MIT FiveK~\cite{MIT_FIVEK:2011} dataset that how an image is exposed through tonal adjustment is person-dependent. \changed{The notion of personalization of image enhancement has also been studied elsewhere (e.g., \cite{Yan:2016, caicedo2011collaborative, SBK10}), which points to its importance.}

\subsection{Challenges}

How would one design a real-time personalized exposure control system? Let us first look at how the control of exposure $EV$ works mathematically. (The definition of exposure is given in the appendix.) Since we bypass the hardware metering function, the exposure value $EV_{i+1}$ for next frame $I_{i+1}$ is fully determined by the current frame $I_i$ (at time $i$). Let us denote $\mathrm{F}(I_i)$ as the function of predicting the exposure adjustment $\Delta^i_{EV}$ given input frame $I_i$, $\mathrm{F}(I_i) = \Delta^i_{EV}$.
The camera will then apply $EV_{i+1} = EV_i + \mathrm{F}(I_i)$ to capture the next frame $I_{i+1}$; this process is iterated in the viewfinder mode\footnote{In practice, the updated $EV_{i+1}$ will not be immediately applied to frame $I_{i+1}$. This is because third-parties typically do not have direct access to the image signal processing (ISP) firmware. Instead, the latency through API calls lasts some number of frames (from our experience, 3 to 5 for iPhone 7 and Google Pixel).}. 

An ideal way of implementing this idea is to collect paired training data ($I_i, \Delta^i_{EV}$) and then perform supervised learning to learn function $\mathrm{F}$. 
Hypothetically, we can do the following: use exposure bracketing to capture various images with different exposure values to create a dataset, then conduct a labeling exercise selecting the best exposure. Subsequently, apply any supervised learning to learn a regression function $\mathrm{F}$ that can map any incorrect exposure to its correct one. However, it is not practical to ask each user to capture and annotate a large scale bracketing dataset in order to train a personalized model. An alternative to acquire training data would be through the ``tap-and-tweak'' mechanism, but again, this approach would not be practical. This is because finding the label $\Delta_{EV}$ associated with the actual corresponding frame in the viewfinder would be non-trivial. Without direct access to the camera hardware (especially for the iPhone), there is a lag when invoking the camera API, which would easily result in mismatched training pairs. All those challenges motivate us to develop an automatic system with a more practical means for acquiring training data.

\subsection{Our Approach}
\label{sec:approach}

\begin{figure}
	\centering{\includegraphics[width=0.8\linewidth]{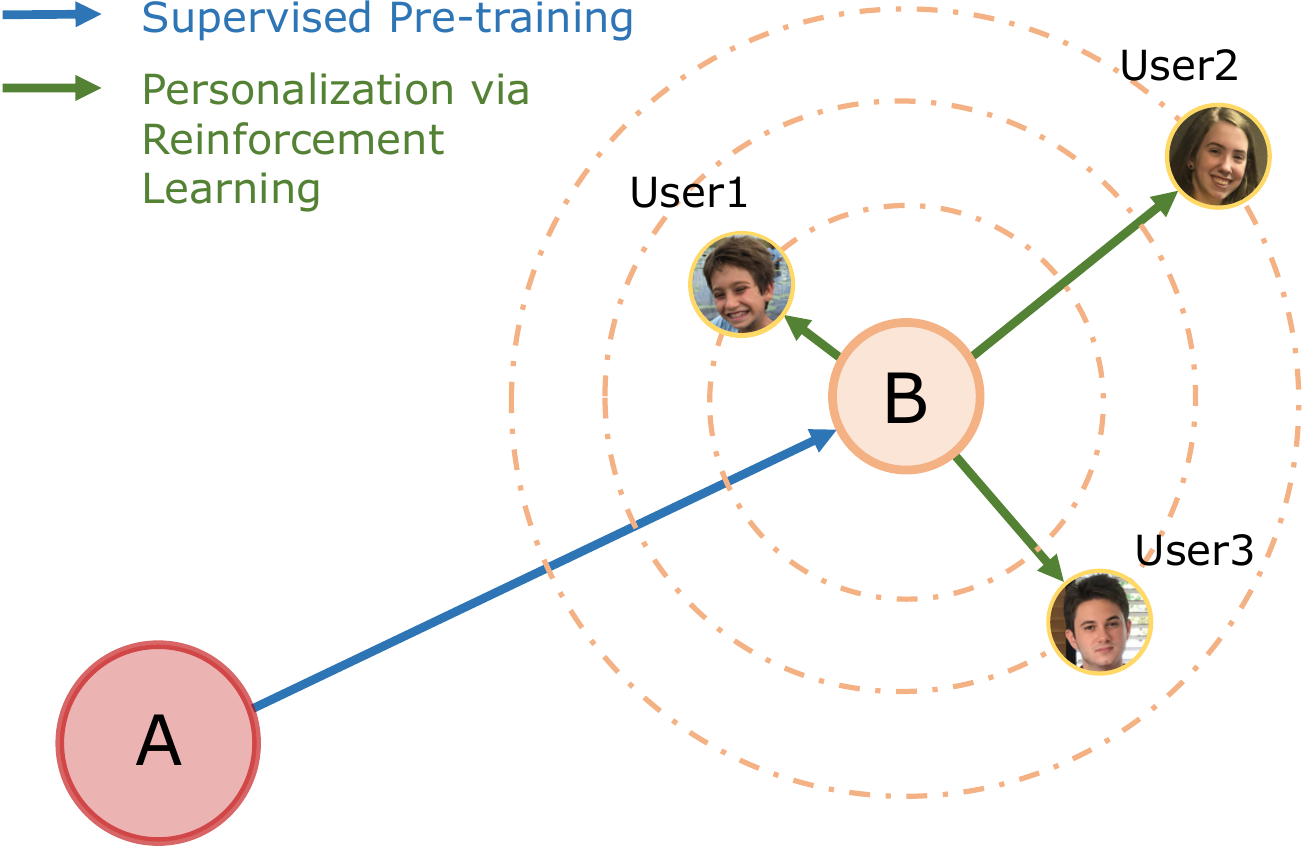}}
	\caption{Our coarse-to-fine learning strategy. At the coarse level, we perform supervised pre-training first to mimic the native camera, which enables us to change from A to B. The fine adjustment is achieved by training the reinforcement learning module on B and learning personalization for different users.}
	\label{fig:keyidea}
\end{figure}

Any well-designed non-native (third party) exposure control system should first perform as well as the native camera for the ``easy'' cases (e.g., scenes with good uniform lighting) while improving on more challenging scenes that require object prioritization (e.g., back-lit scenes). This inspires us to adopt a ``coarse-to-fine" learning strategy, with ``coarse" being pre-training using native camera data and ``fine" being fine-tuning using on-line learning. This strategy is illustrated in Figure~\ref{fig:keyidea}.

To enable semantic understanding, we use a fully convolutional neural (FCN) network to represent the exposure prediction function $F$. The FCN network is pre-trained through supervised learning on a synthetic dataset. After the pre-training, our model mimics the behavior of the native camera and can be deployed to each end-user. We call this model as the basic or average model. Once deployed, the basic model is then fine-tuned locally through the on-line learning module. 

During the on-line stage, at time $t$, the hardware can only choose one specific $EV$ and capture a corresponding image $I$, which is unlike full exposure bracketing. Note that it is impractical to ask users to directly provide the annotation of $\Delta_{EV}$ for image $I$ without the corresponding exposure bracketing set. However, the user could instead provide feedback on the exposure after capture, namely, if the captured image is ``under-exposed," ``correctly exposed," or ``over-exposed." 

Such feedback serves as a reward signal to indirectly supervise how the system intelligently selects $\Delta_{EV}$ for a given image. This is where reinforcement learning comes in; it allows both data collection and personalization to be feasible and scalable. After we accumulate a new batch of images with their corresponding reward signals, the local model is then fine-tuned through back-propagation on this new batch based on a Gaussian policy gradient method. This process is iterated until all the feedback signals are positive. 

Most native cameras have default metering options for auto-exposure. Each option works by assigning a relative weighting to different spatial region. However, such weighting schemes are all heuristically pre-defined, such as spot or center-weighted metering. (Newer cameras also use face information for exposure prioritization.) To generalize metering and improve the learning performance, we introduce an \changed{adaptive metering} module into the FCN network. For each image frame, the \changed{adaptive metering} module outputs a weighting map which is element-wise multiplied by another learned exposure map. The whole system is learned end-to-end. In this paper, we show the effectiveness of the \changed{adaptive metering} module both visually and quantitatively. Once the model is trained, during run-time, we directly feed-forward the current frame into the network to get the output $\Delta_{EV}$, which then used to capture the next frame.



Our major contributions are as follow:
\begin{itemize}
	\item To the best of our knowledge, our work is the first to address personalized real-time exposure control based on learned scene semantics.
	\item We propose a practical ``coarse-to-fine" learning strategy which first uses supervised learning to achieve a good anchor point, followed by refinement through reinforcement learning. We believe that such a training strategy could inspire similar approaches to other problems in computer vision. 
    \item We introduce an \changed{adaptive metering} module that can automatically infer user prioritization in a scene. We show that it improves the learning performance.
	\item We develop an end-to-end system and implement it on iPhone 7 and Google Pixel and demonstrate good performance in terms of both speed and visual quality. Our system learns the proper exposure for a variety of scenes and lighting conditions, and outperforms its native camera counterparts for challenging cases that require general semantic understanding.
\end{itemize}

\begin{figure*}[!t]
    \centering
     \includegraphics[width=\linewidth]{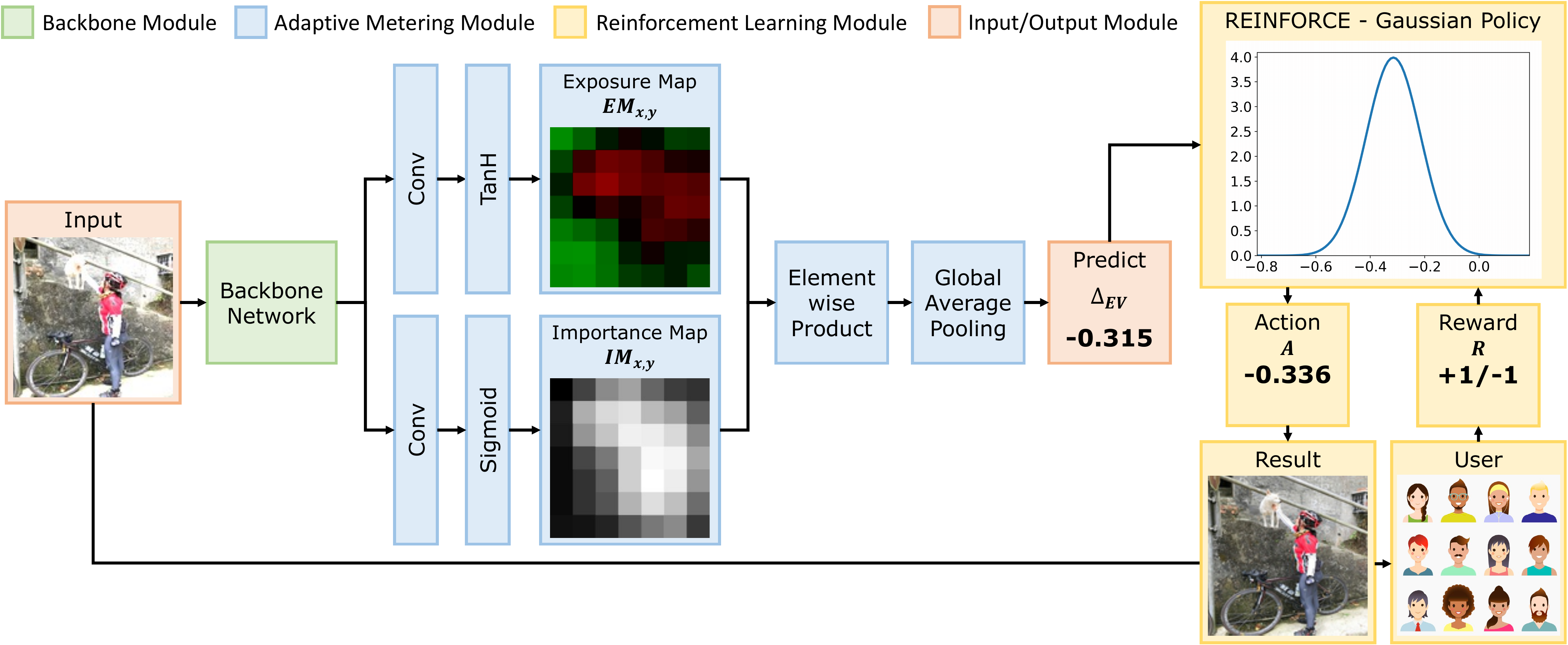}
	\caption{Overview of our personalized auto-exposure framework. The backbone network is designed as a light-weight deep convolutional neural network which trades-off between accuracy and performance on mobile devices. The \changed{adaptive metering} module (which contains the \changed{importance map $IM$} and exposure map $EM$) learns prioritization of exposure setting based on scene content. The reinforcement module uses policy gradient with Gaussian policy for online learning; the reward is based on user feedback. Note that in exposure map, we use green, red and black to represent positive, negative and zero $\Delta_{EV}$ respectively.}
    \label{fig:overview}
\end{figure*}

\section{Related Work}

In this section, we briefly review prior work on auto-exposure and post-processing techniques for tonal enhancement.

\subsection{Auto-Exposure}

Except a few early papers (e.g., \cite{brunner2012automatic,AutoExposure_2002}), there appears to be little published research work on exposure control for mobile cameras. For competitive reasons, camera manufacturers do not usually publicize full details of their technique for real-time auto-exposure, though information for a few camera models exists (see, for example, \cite{Sampat:1999}). In native mode, cameras rely on built-in metering to measure the amount of light for determining the proper exposure values. This works well in many cases. However, assuming different options for meter mode are available, users may need to modify the meter mode (e.g., spot, matrix) and focus area for handling more challenging conditions such as back-lit scenes. Techniques for more specialized treatment of front- and back-lit scenes have been described (e.g., \cite{Guo:2014}), but they fundamentally rely on histogram analysis that does not jointly consider object importance and lighting condition. 
Some native cameras on popular phones prioritize exposure on detected face regions~\cite{Face_exposure_1,Auto_exposure_roi17,Yang_facedetection}; many allow user interaction to select a region of interest for appropriate exposure adjustment. In contrast, our auto-exposure approach supports more general scenes beyond faces, and it does not require user interaction. 

\subsection{Post-Processing Tonal Adjustment}

There are many post-processing techniques for enhancing image quality (e.g., \cite{SBK10,MIT_FIVEK:2011,Joshi:2010,BD:2016,Berthouzoz:2011:FCP,Kaufman:2012,Lischinski:2006:ILA,lou2015color, guo2010correcting, xu2011correction, eilertsen2017hdr, endo2017deep}) or achieve certain artistic effects, (e.g., \cite{Wang:2011:EIC,HaCohen:2011:NDC,Gatys_2016_CVPR,Yan:2016}). 
We believe that optimizing exposure during capture is valuable as it captures details of important objects; once they are lost due to inappropriate exposure, no amount of post-processing would help to recover the original information. More recently, \cite{Exposure:hu} introduced a new framework for photo retouching that tries to model and learn the photo editing operations using both adversarial and reinforcement learning. Although they also model the exposure adjustment in addition to color and contrast, the problem setting is significant different from ours and their goal is not for real-time personalized exposure control. Table~\ref{tab:intro_compare} compares three features (ability to be scene-aware, personalized, and real-time for mobile devices) for post-processing techniques, the native camera, and our system. 

\subsection{High Dynamic Range (HDR) Imagery}

The system reported in \cite{Hasinoff:2016} optimizes image quality through raw burst capture and efficient image processing. While it has an example-based auto-exposure component along the pipeline, the goal is to deliberately underexpose the image to better tailor the subsequent HDR+ fusion. In addition, Hasinoff et al. use low-level features and nearest neighbor search to account for semantics and weighted blending of exposure values from matched examples, whereas we use an end-to-end deep learning system for semantic understanding related to optimal exposure values. A more recent paper \cite{gharbi2017deep} proposes a lightweight CNN approach to regress the affine transforms in the bilateral space in order to create edge-preserving filters for real-time image enhancement. While their results look impressive, their system requires pair-wise training data to perform fully supervised learning. By comparison, we apply reinforcement learning due to the lack of directly label data.

Compared with the above approaches, another major differentiator is our work can be treated as a solution to a control problem; we not only require the steady-state exposure value to be optimal, but we also want fast convergence without oscillatory behavior for different scenes and lighting conditions in the viewfinder. 

\begin{table}[!h]
\centering
\begin{tabular}{cccc}
\toprule
 & Post-processing & Native & Ours \\
\midrule
Scene-Aware & Partially$^{(a)}$ & Partially$^{(b)}$ & Yes\\
Personalized & Partially$^{(a)}$ & No & Yes\\
Real-Time$^{(c)}$ & No & Yes & Yes\\
\bottomrule
\end{tabular}
\caption{Feature comparison between post-processing approaches, native approach, and our approach. Notes: 
$(a)$ Many techniques are not content-aware and/or personalized. 
$(b)$ Exposure prioritization is based on faces and possibly very simple scenes, such as sky \cite{Kaufman:2012}.
$(c)$ Real-time on mobile devices.}
\label{tab:intro_compare}
\end{table}

%% file: overview.tex

\section{System Overview}

Our system is depicted in Figure~\ref{fig:overview}. It is based on a fully convolutional neural network.

\subsection{Network Structure}

Apart from the input and output, our system consists of three sequentially connected components: (1) backbone network (within the green box), (2) \changed{adaptive metering} module (within the blue box), and (3) reinforcement module (within the yellow box).

\subsubsection{Backbone Module}
A backbone network can be any network structure as long as it extracts semantically meaningful representations. In the context of real-time exposure control, the backbone network should be designed for both accuracy and run-time performance. Recent studies on image classification (e.g., \cite{howard2017mobilenets,darknet13,SqueezeNet}) show that the backbone could be significantly reduced in terms of the number of layers and model size without losing accuracy. In addition, the state-of-the-art system~\cite{li2017light} for object detection shows that when the head is carefully designed, the backbone does not have to be very deep.

\subsubsection{Adaptive Metering Module}
\label{sec:adaptivemetermodule}

\changed{To predict the important regions for exposure prioritization,} we introduce an \changed{adaptive metering} module as a means for semantic-based metering. (The adaptive metering module sits on top of the backbone net.) The goal of this module is to learn both an exposure map (denoted as $EM_{xy}$) and an \changed{importance map (denoted as $IM_{xy}$)} in parallel. The exposure and importance maps are element-wise multiplied and then processed through a global average pooling layer to output the final $\Delta_{EV}$ (Figure~\ref{fig:overview}).

It is possible to train a general neural network to learn $\Delta_{EV}$ implicitly. However, using a general version may result in convergence and performance issues. \changed{Our network design with split EM and IM branches has good convergence behavior. This design is also intuitive in that it mimics} how exposure is typically computed in current camera systems (namely, prioritizing image areas and computing overall exposure based on the resulting weight distribution). Our network is trained to prioritize based on scene.

\subsubsection{Reinforcement Learning Module}
\changed{As described in Section 1, it is not practical to ask each user to capture and annotate a large scale bracketing dataset in order to train our system.} Instead, it is easier to ask user to label the exposure quality in one of the three categories: ``over-exposed'',``under-exposed'' and ``well-exposed''.  So, if an image is labeled as ``well-exposed", the output delta EV should be zero; when an image is labeled as ``under-exposed" (``over-exposed"), the output delta EV should be positive (negative). Such label information is incomplete and weak, since it does not provide the exact exposure value for a given photo. To overcome these challenges, we propose a third module that is based on reinforcement learning. During run-time, the reinforcement module would no longer be required. We can directly feed the current frame into the network and output the corresponding $\Delta_{EV}$, which is then used to capture the next frame.



\subsection{Training Strategy}

To apply reinforcement learning to our personalized exposure control system, we have to define the state, action, and reward. Intuitively, the state represents any frame in the viewfinder, regardless their exposure quality, while the action is defined as any $\Delta_{EV}$ that lies in the range of possible exposure adjustments. We use a 1D Gaussian parameterized by mean $\mu$ and a constant standard deviation $\sigma$ to sample an action during the training stage. It is reasonable to ask users to provide feedback on exposure quality through a simple selection from over-, under- or well-exposed options. Once acquired, such data are directly used as the reward signal to train the whole system. We believe that this design choice facilitates on-line learning to personalize the exposure model for each user or even each device. 

Unfortunately, directly training from scratch would require significant amounts of data and thus result in slow convergence; this is especially cumbersome to generate a personalized model. Instead, we first train the backbone network with an \changed{adaptive metering} module via supervised learning to replicate a basic exposure control behavior. Since such a pre-trained network can already give us a good initial estimate $\Delta_{EV}$, we only need to fine-tune the reinforcement module by locally refining $\Delta_{EV}$ based on the reward signals (as shown in Figure \ref{fig:keyidea}). This significantly simplifies learning while being more likely to end up with a better local minimum. In addition, since fine-tuning is done in an end-to-end training fashion, the reinforcement learning module can more easily customize the learned semantic features embedded in all the layers to handle exposure control. More training is expected to improve robustness and accuracy of the control policy.

%% file: attention.tex
\section{Exposure Control}
\label{sec:attention}

Unlike post-processing techniques that are applied to photos {\em after} they have been captured, our goal is real-time exposure control in the viewfinder mode. In addition to generating correct steady-state exposure values, our system also needs to take into account time performance, convergence speed, and avoidance of oscillatory behavior. 

\subsection{Exposure-Control As Markov Decision Process}

To simplify the model and reduce the computational overhead, we assume that the exposure control process in the viewfinder mode can be treated as a Markov Decision Process (MDP): the new state $s^t$ (next captured frame $I_{i+1}$) is determined by the current state $s$ (current frame $I_i$) and action $a$ (exposure adjustment $\Delta_{EV}$). In other words, given $s$ and $a$, the new state is assumed to be independent of previous states and actions. Therefore, in order to achieve fast convergence and reduce the observation latency in the viewfinder, we want the next captured frame to be as close as possible to the optimal one given the current frame, which requires the exposure adjustment $\Delta_{EV}$ to be directly close to optimal one as well. If $\Delta_{EV}$ is too large, it may result in overshooting and therefore causing oscillation, while a value that is too small would result in slow convergence and large latency. 

We want to learn an exposure prediction function $F(I_i)$ to generate the optimal exposure adjustment $\Delta_{EV}$. As mentioned earlier, there is no scalable way to collect training pairs ($I, \Delta_{EV}$) for supervised learning. To overcome this problem, we use reinforcement learning with a Gaussian policy to learn $F(I_i)$. One advantage of our approach is the ease with which the exposure control model can be personalized for each user. More details are given in Section~\ref{sec:reinforce}.

\subsection{Adaptive Metering}

Smart phone cameras generally rely on firmware metering modes (spot, matrix, center weighted, and global) to determine the exposure value. Newer high-end cameras are capable of detecting faces and possibly categorizing simple scenes, and such information is used to optimize exposure. However, to our best knowledge, there are no technical details on how these cameras make use of scene semantics for automatic real-time auto-exposure.

To adjust the exposure, users are given the option to manually tap on the screen to select the region of interest. There is also the option to additionally adjust the luminance (Figure~\ref{fig:iphone_capture}). Given the manual nature of the exposure change, it would be easy to miss good moments to capture (e.g., when capturing a fast moving object of interest such as a bird in flight). It is much more desirable to have the camera automatically produce the optimal exposure at all times---this is the goal of our work.

The hardware metering modes are heuristic approaches for exposure control. For better generalization (beyond faces), we need a systematic approach to adaptively predict a \changed{metering importance} map based on the scene. Our \changed{importance} map is a normalized weighting map that is used to display the importance of each spatial region (e.g., car, building, pet, and flower) as learned through examples.  \changed{Therefore, we propose to develop an adaptive metering module which consists of the exposure adjustment map (how exposure should change locally) and the weighting map (spatial distribution of importance).}

\changed{These two maps are computed using two different activation functions, namely, \textbf{TanH} and \textbf{Sigmoid}. The feature map computed using \textbf{TanH} is interpreted as the exposure map (EM). Each value (within $[-1.0, 1.0]$) represents the amount of exposure, with positive values indicating over-exposure and negative values indicating under-exposure. On the other hand, the feature map computed using \textbf{Sigmoid} is interpreted as the importance map (IM). Each value (within $[0.0, 1.0]$) represents the importance for determining exposure. The exposure and importance maps are element-wise multiplied and then processed through a global average pooling layer to output the final $\Delta_{EV}$:}

\begin{equation}\label{equ:final_delta_ev}
\Delta_{EV} = \frac{1}{HW}\sum_{x=1}^{H}\sum_{y=1}^{W}EM_{x,y}\times \changed{IM_{x,y}} .
\end{equation}

Suppose we have the ground-truth exposure adjustment $Y_i$ for each image $I_i$. Training the exposure prediction function $F(I_i)$ is straightforward, by minimizing the following loss: 
\begin{equation}\label{equ:attention_loss}
\mathcal{L} = \frac{1}{N}\sum_{i=1}^{N}(F(I_i) - Y_{i})^2 .
\end{equation}

Because we use a deep convolution neural net to represent function $F$, minimizing loss $\mathcal{L}$ is done simply through back propagation. However, in practice, as we discussed before, there is no ground-truth for each image. In this paper, we propose to use reinforcement learning with a Gaussian policy to learn the function $F$ (Section~\ref{sec:reinforce}). In Section~\ref{sec:experiment}, we show examples to illustrate that our system learns semantic information for adaptive metering.


Once the model is deployed, during run-time, our exposure control system will be used for capturing a new frame during the viewfinder mode. The moment the viewfinder is turned on, the hardware captures the first frame denoted as $I_0$, which is then fed into our exposure prediction network to obtain $\Delta_{EV_{0}}$. This is added to $EV_{0}$ to generate the new exposure value $\Delta_{EV_{1}}$, which is passed to the hardware control system to capture the new frame. If the predicted $\Delta_{EV}$ drops to zero, it means the exposure has converged to a steady state (assuming static camera and scene). 

Ideally, the steady state is achieved in one step. In practice, this is not possible due to firmware latency, especially when there is no direct control over the firmware. Empirically, even though we train our system to directly predict the optimal exposure adjustment, it takes 3 to 5 frames for convergence. 
%
%

\begin{figure}[!t]
	\centering
	\includegraphics[width=\linewidth]{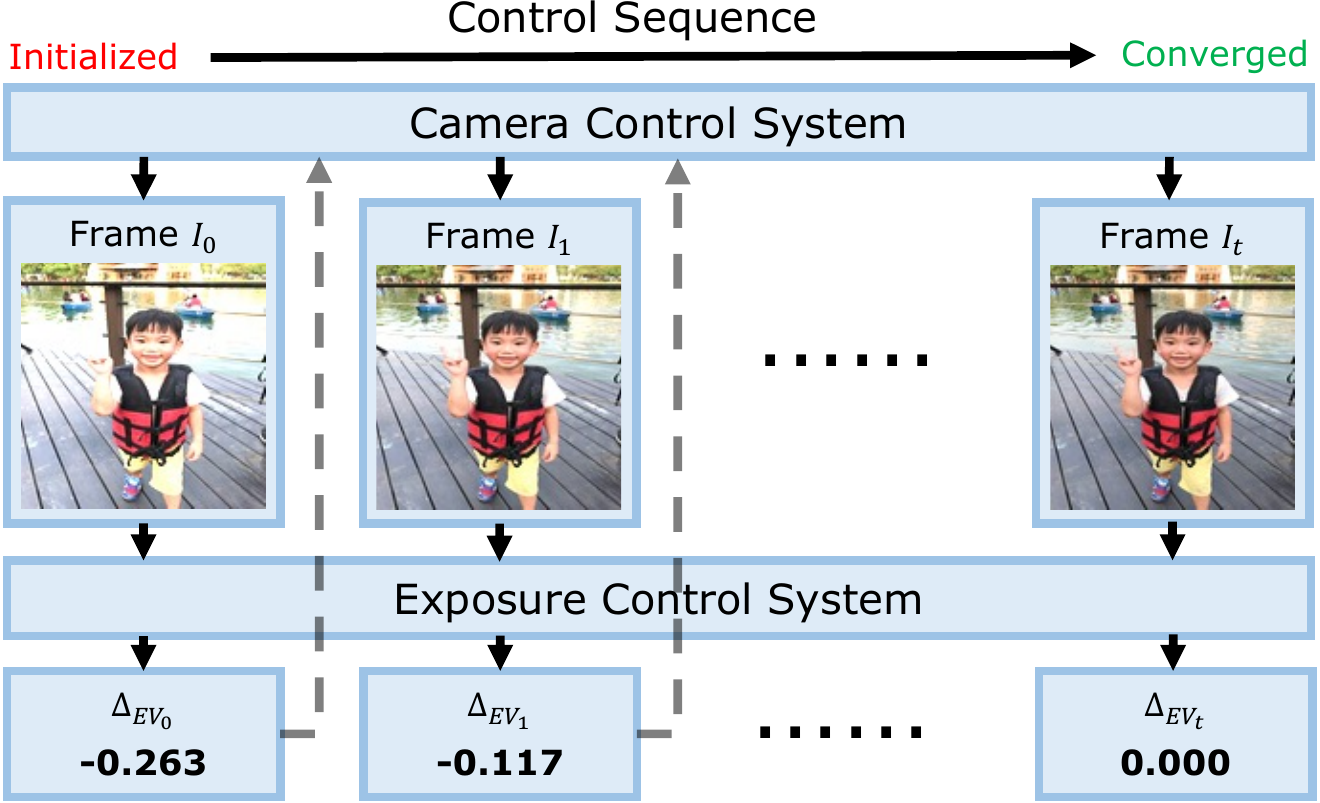}
	\caption{Flow of our exposure control system during run-time.}
	\label{fig:stage1}
\end{figure}

%% file: reinforce.tex
\section{Reinforcement Learning} \label{sec:reinforce}


 

As discussed earlier, there is no scalable way to easily obtain pair-wise training data (consisting of $(I,\Delta_{EV})$ pairs) for supervised learning. As can be seen in the MIT FiveK~\cite{MIT_FIVEK:2011} dataset, exposure and tone adjustment are highly person-dependent. In addition, given differences in the camera firmware and ISP (image signal processing), the temporal response and steady-state appearance vary from one device to another. In practice, it is rare for third parties to have direct control of the camera hardware, and the detailed operations of the ISP (which likely involve nonlinear mappings) are mostly unknown. 

Despite these difficulties, we believe it is better to train using processed RGB images rather than raw (Bayer filtered) images. This is because we wish \textit{the final processed images to achieve the correct exposure}, i.e., images generated right after passing through the ISP pipeline. Training using raw images would be problematic because the camera firmware and ISP are typically black boxes; what may work for one camera would likely not work as well for another. We cast exposure control as an MDP (Markov Decision Process) problem and use reinforcement learning to train the system.

\subsection{Formulation}
In the context of reinforcement learning, an image is a state, with $\mathcal{S}$ representing the set of \textbf{states}. $\mathcal{A}$ is the set of possible \textbf{actions}, where each action is one specific exposure adjustment value $\Delta_{EV}$. The \textbf{environment} consists of two components, namely, camera and user. The user responses are used to provide the \textbf{reward} to the image captured by camera while the camera continually receives new exposure adjustment commands. The \textbf{agent} is the fully convolutional neural network shown in Figure~\ref{fig:overview}. 

\begin{figure}[!t]
    \centering
    \includegraphics[width=\linewidth]{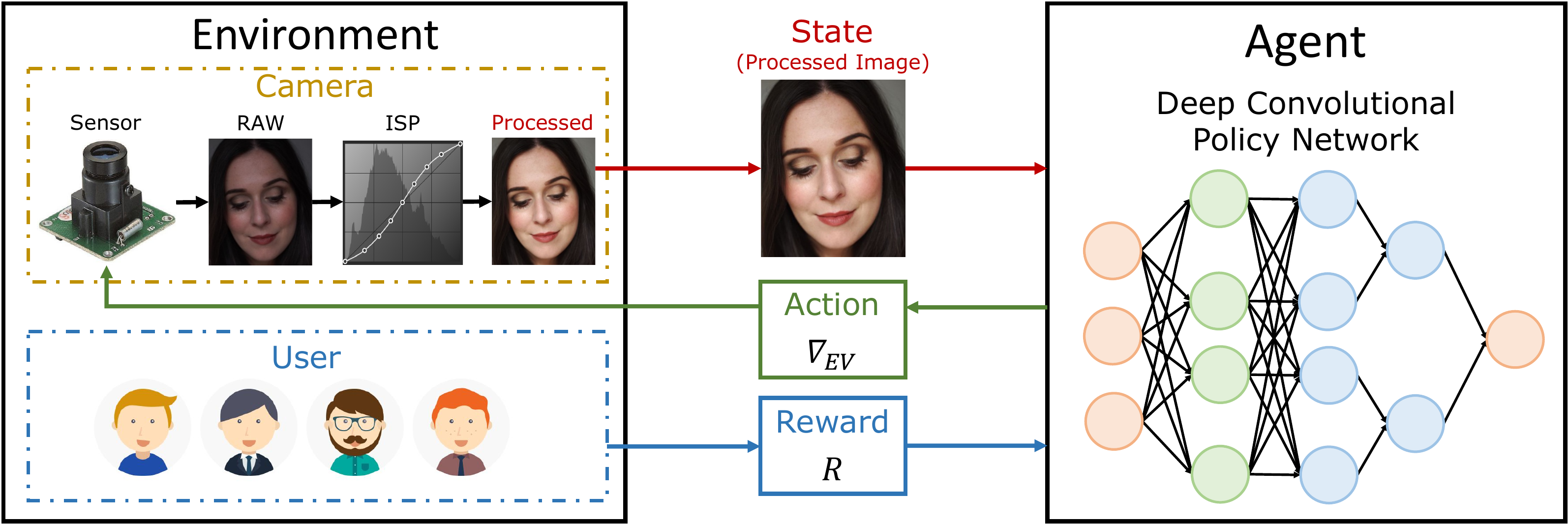}
    \caption{The MDP components in our control system. }
    \label{fig:rldef}
\end{figure}

Figure~\ref{fig:rldef} shows the components in our MDP framework. The goal of the agent is to refine its control policy to adapt to different users and camera firmware based on the rewards. Therefore, even though the camera firmware and ISP are collectively treated as a single black box, we can generate good camera exposure control behavior indirectly through the learned policy network. 

At each time step $t$, the camera acquires an image $s_t$, which is then judged by a user as being over-exposed, under-exposed, or well-exposed. (Section~\ref{sec:reward} provides more details on how such feedback is converted to reward used to train the whole system.) The database of image-reward pairs is used to then update the policy network through a policy-gradient algorithm with Gaussian policy. We discuss the policy-gradient algorithm in Section~\ref{sec:policy_gradient}. While not currently demonstrated in this work, our formulation is general enough to customize device behavior.


\subsection{Policy Gradient with Gaussian Policy}
\label{sec:policy_gradient}

To train the system, we choose policy gradient rather than Q-learning for two reasons. One reason is policy gradient generally performs better than Q-learning on a large action space. The other reason is policy gradient supports continuous action space while Q-learning usually requires discrete actions. 

As a comparison, we partition the exposure adjustment values into small discrete values and then use classification to train the system. However, we encounter the problem of temporal oscillatory behavior. As a result, we choose to directly regress the exposure adjustment value $\Delta_{EV}$ as a continuous variable, with values closer to the optimal being penalized less than those farther away. Based on this observation, we use a Gaussian to serve as the policy function to sample each action during training.

Let $\pi_{\theta}$ (parametrized by $\theta$) be the policy function and $J(\theta)$ the expected total reward; the goal is to maximize $J(\theta)$:
\begin{equation}\label{equ:rl_obj}
\begin{array}{lll}
J(\theta)&=&\mathbb{E}_{\pi_{\theta}}[r] \\
         &=&\sum\limits_{s\in\mathcal{S}}d(s)\sum\limits_{a\in\mathcal{A}}\pi_{\theta}(a|s)r(s,a) ,
\end{array}
\end{equation}
where $d(s)$ is the probability distribution for each state, $\pi_{\theta}(a|s)$ is the conditional probability distribution of each possible action $a$ given the state $s$, and $r(s,a)$ is the reward after applying action $a$ given state $s$.

Following REINFORCE~\cite{Williams1992}, we compute the derivative of the objective function $J(\theta)$ with respect to the parameters $\theta$ as 
\begin{equation}\label{equ:rl_obj_grad_1}
\begin{array}{lll}
\nabla_{\theta} J(\theta)&=&\nabla_{\theta}\sum\limits_{s\in\mathcal{S}}d(s)\sum\limits_{a\in\mathcal{A}}\pi_{\theta}(a|s)r(s,a)\\
                         &=&\sum\limits_{s\in\mathcal{S}}d(s)\sum\limits_{a\in\mathcal{A}}\nabla_{\theta}\pi_{\theta}(a|s)r(s,a).
\end{array}
\end{equation}
With
\begin{equation}\label{equ:rl_obj_grad_trick}
\nabla_{\theta}\pi_{\theta}(a|s)=\pi_{\theta}(a|s)\nabla_{\theta}\log\pi_{\theta}(a|s),
\end{equation}
Eq.~\ref{equ:rl_obj_grad_1} can rewritten as
\begin{equation}\label{equ:rl_obj_grad_2}
\begin{array}{lll}
\nabla_{\theta}J(\theta)&=&\sum\limits_{s\in\mathcal{S}}d(s)\sum\limits_{a\in\mathcal{A}}\pi_{\theta}(a|s)\nabla_{\theta}\log\pi_{\theta}(a|s)r(s,a)\\
                         &=&\mathbb{E}_{\pi_{\theta}}[\nabla_{\theta}\log\pi_{\theta}(a|s)r(s,a)].
\end{array}
\end{equation}


As mentioned before, we use a Gaussian function to model the policy. As such, $\pi_{\theta}$ can be formulated as
\begin{equation}\label{equ:rl_gp_1}
\pi_{\theta}=\mathcal{N}(\mu(s); \sigma^{2})=\mathcal{N}(F_{\theta}(s); \Sigma),
\end{equation}
where $F_{\theta}$ is the parameterized fully convolutional network that outputs $\Delta_{EV}$ for any input image, as discussed in Section~\ref{sec:attention}. For simplicity, we directly take the network output $\Delta_{EV}$ as the mean value $\mu(s)$ of our Gaussian policy function, and take a constant value as its variance $\Sigma$.  We empirically set $\Sigma$ to $0.1$ in our current experiments. During training, we sample the action based on this Gaussian policy function, while at run-time, we directly use the output as the final action to control the exposure adjustments.

Based on Gaussian policy, $\log\pi_{\theta}(a|s)$ in Eq.~\ref{equ:rl_obj_grad_2} can be rewritten as 
\begin{equation}\label{equ:rl_gp_2}
\log\pi_{\theta}(a|s)=-\frac{1}{2}||F_{\theta}(s)-a||^{2}_{\Sigma^{-1}}+\text{const},
\end{equation}
where $||F_{\theta}(s)-a||^{2}_{\Sigma^{-1}}=(F_{\theta}(s)-a)^{T}\Sigma^{-1}(F_{\theta}(s)-a)$. 

Then the derivative of $\log\pi_{\theta}(a|s)$ in Eq.~\ref{equ:rl_obj_grad_2} can be further rewritten as
\begin{equation}\label{equ:rl_gp_3}
\nabla_{\theta}\log\pi_{\theta}(a|s)=-\Sigma^{-1}(F_{\theta}(s)-a)\nabla_{\theta}F_{\theta}(s).
\end{equation}
As a result, the objective function in Eq.~\ref{equ:rl_obj_grad_2} is rewritten as
\begin{equation}\label{equ:rl_gp_4}
\nabla_{\theta}J(\theta)=\mathbb{E}_{\pi_{\theta}}[-r(s,a)\Sigma^{-1}(F_{\theta}(s)-a)\nabla_{\theta}F_{\theta}(s)].
\end{equation}

To estimate $\nabla_{\theta}J(\theta)$ in Eq.~\ref{equ:rl_gp_4}, we sample over the action space using the probability distribution $\pi_{\theta}(a|s)$:
\begin{equation}\label{equ:rl_obj_grad_3}
\nabla_{\theta}J(\theta)\approx -\frac{1}{N}\sum\limits_{i=1}^{N}r(s_i,a_i)\Sigma^{-1}(F_{\theta}(s_i)-a_i)\nabla_{\theta}F_{\theta}(s_i).
\end{equation}
We describe the reward function $r(s_i,a_i)$ in the next section.

\subsection{Reward Setting and Supervised Pre-training}
\label{sec:reward}

Earlier, we discussed that it is easier to collect a database of images and their corresponding user feedback on exposure quality (over-exposed, under-exposed, or well-exposed). However, we still need to address two challenges before applying reinforcement learning. 

The first challenge is that once our basic system is deployed, before any personalization learning is conducted, we can only collect the feedback signals for those images directly captured by our system. More specifically, in the context of reinforcement learning, for each state image $I$, we can only obtain the feedback signal for action $A$, where $A=\Delta_{EV}=0$. This is because, our exposure control system is generally considered to have converged (i.e., $\Delta_{EV}=0$) before shutter is pressed. 

So far, what we can use to train are only the state-action-reward tuples $(I, A=\Delta_{EV}=0, R)$. During the personalization training stage, our Gaussian policy requires sampling of different actions $A\neq 0$ for state image $I$. The second challenge is how to get the corresponding rewards, given that it is not practical to require the user to provide such feedback. 

\begin{figure}[t!]
    \centering
    \includegraphics[width=\linewidth]{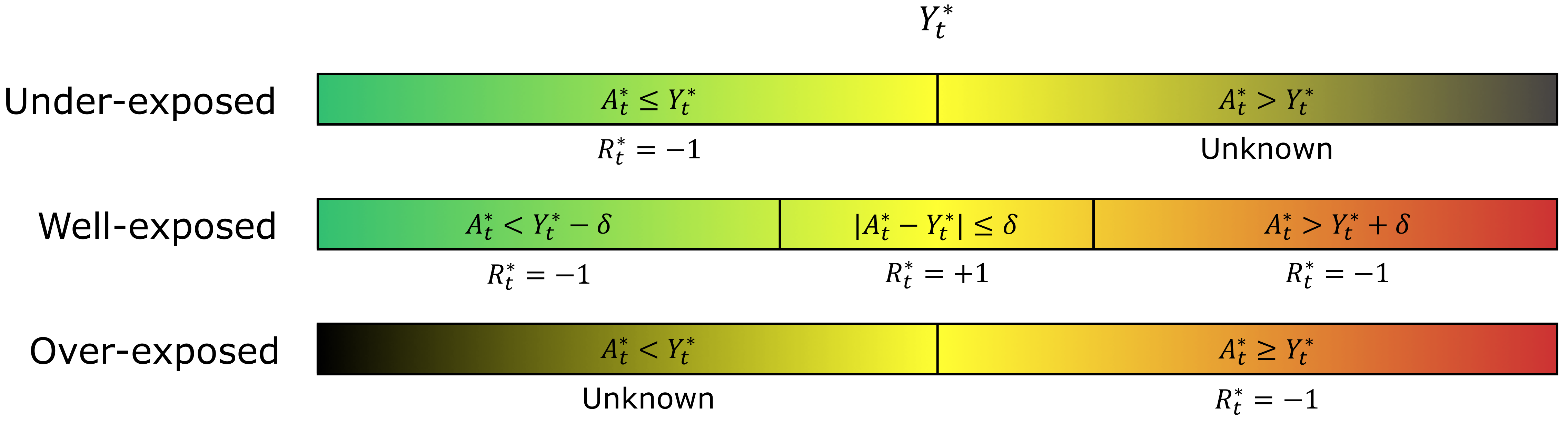}
    \caption{Strategy to compute reward, with $\delta=0.1$.}
    \label{fig:rlrwd}
\end{figure}

To augment our training set, we synthesize a set of new images from each original image $I$ corresponding to different exposures. Prior to synthesis, we linearize the image intensities by applying a power curve (with $\gamma=2.2$)\footnote{Making the image linear is necessary to simulate changes in exposure.}. We use Adobe Lightroom to synthesize new images $I_t$ from linearized images by synthetically changing $\Delta_{EV}$ from $-2.0EV$ to $2.0EV$ with a step of $0.2EV$; note that we reintegrate the nonlinearity to these synthetic images using the power curve with $\gamma = 1.0/2.2$. Each new synthesized image $I_t$ corresponds to one action $Y_t^{*}= -2.0EV+t*0.2EV$, i.e., represents the consequence of applying that action to the original image $I$. For any image $I_t$ (synthetic or real), we need to define its reward $R^*_t$ based on both $Y_t^{*}$ and the original feedback signals for $I$ (where $I=I_{10}$). Figure ~\ref{fig:rlrwd} illustrates how to compute the reward $R^*_t$ for any sampled action $A^*_t$.

Our reward system is based on the intuition that for any ``well-exposed'' image, a small perturbation of the exposure (i.e., small $\Delta_{EV}$) is acceptable. Eventually, as the magnitude of $\Delta_{EV}$ increases, the image will become either ``under-exposed'' or ``over-exposed''. For the case that the image is ``under-exposed'', if we further decrease the exposure will obviously result a more unacceptably ``under-exposed'' image. The logic can be similarly extended to the ``over-exposure'' case.

In our system, if the original image is annotated as ``well-exposed'', then for any other synthetic image $I_t$ and any sampled action $A_t^*$, as long as $A_t^*$ is very close to $Y_t^*$, the reward $R_t^*$ should be positive which is shown as the middle region in the second row in Figure ~\ref{fig:rlrwd}. Conversely, if the sampled action is far away from $Y_t^*$, we should penalize that action through a negative reward (corresponding to the left and right regions in the second row in Figure ~\ref{fig:rlrwd}).

If the original image is annotated as ``under-exposed'', any sampled action $A_t^*$ decreasing its exposure should make it even more ``under-exposed''. In this case, the reward $R_t^*$ should be negative, which corresponds to the left region in the first row in Figure ~\ref{fig:rlrwd}. However, for the same ``under-exposure'' case, it is uncertain as to what the reward should be if the sampled action $A_t^*$ is unexpectedly larger than $Y_t^*$ without further supervision. We ignore such cases in the training which is shown as unknown region in the first row in Figure~\ref{fig:rlrwd}. (The same situation applies when we deal with images annotated as ``over-exposed'' which is shown in the third row in Figure ~\ref{fig:rlrwd}.)

Reinforcement learning generally needs lots of data and long convergence time for it to be effective. As shown in Figure~\ref{fig:keyidea}, we propose to pre-train the network using supervised learning. On one hand, it is reasonable to assume that any personalized exposure control system should at least perform as well as a native camera; on the other hand, from a data-collection standpoint, supervised pre-training for mimicking native camera can be conducted easily on a relative large scale set, which does not require human annotation.

%% file: experiment.tex
\section{Datasets}

In Section~\ref{sec:experiment}, we show experimental results to demonstrate the effectiveness of supervised pre-training and our \changed{adaptive metering} module as well as the performance of reinforcement learning for personalized exposure control. The results are based on four different datasets, namely, Flickr, MIT FiveK, iPhone 7, and Google Pixel datasets. Table~\ref{tab:datasets} shows the number of images in each dataset. We now describe how each dataset is used.

\begin{table}[h!]
\centering
\begin{tabular}{ccccc}
\toprule
\multicolumn{1}{l}{} & \multicolumn{1}{l}{Flickr} & \multicolumn{1}{l}{MIT FiveK \cite{MIT_FIVEK:2011}} & \multicolumn{1}{l}{iPhone 7} & Google Pixel \\ \midrule
\#Images & 11,000 & 5,000 & 26,679 & 7,056 \\
Format & JPEG & RAW & JPEG & JPEG \\
\bottomrule
\end{tabular}
\caption{Datasets used in our experiments.}
\label{tab:datasets}
\end{table}


Each dataset is used for a different purpose: Flickr dataset for supervised pre-training, MIT FiveK dataset for personalization, and iPhone 7 and Google Pixel datasets for enhancement of native camera performance of the respective smartphones.

\subsection{Flickr Data For Supervised Pre-training}

Our system requires a dataset that can be used for supervised pre-training in order to mimic the exposure control behavior of native cameras. To this end, we downloaded 100,000 images that were captured by mobile devices from Flickr.
To improve image diversity, we remove duplicates and then randomly sample 11,000 of them to serve as the original dataset for supervised pre-training. As discussed in Section~\ref{sec:reinforce}, to further augment the training set, for each image, we use Lightroom to synthesize 20 images corresponding to $\Delta EV$ from $-2.0 EV$ to $+2.0 EV$ at a step of $0.2 EV$. As a result, we end up with a large pair-wise training set, each of which contains a synthetic image and its corresponding $\Delta_{EV}$.

The supervised pre-training is straightforward, where the loss function is simply the Euclidean loss. We simplify pre-training by assuming the original image to be well-exposed. Please note that the upper bound performance of the pre-training model matches that of the native camera. In this work, the pre-training model is just a starting point that is to be enhanced, in order to exceed the performance of the native camera app and adapt to personalized preferences.


\subsection{MIT FiveK Dataset for Personalization}

One approach to test the performance of reinforcement learning for personalization would be to deploy our basic model (after supervised pre-training) to the user's device first and then capture and annotate many images as being under-exposed, well-exposed, or over-exposed. Instead, we choose to leverage the MIT FiveK dataset \cite{MIT_FIVEK:2011}.

\begin{figure}[h!t]
    \centering
    \includegraphics[width=\linewidth]{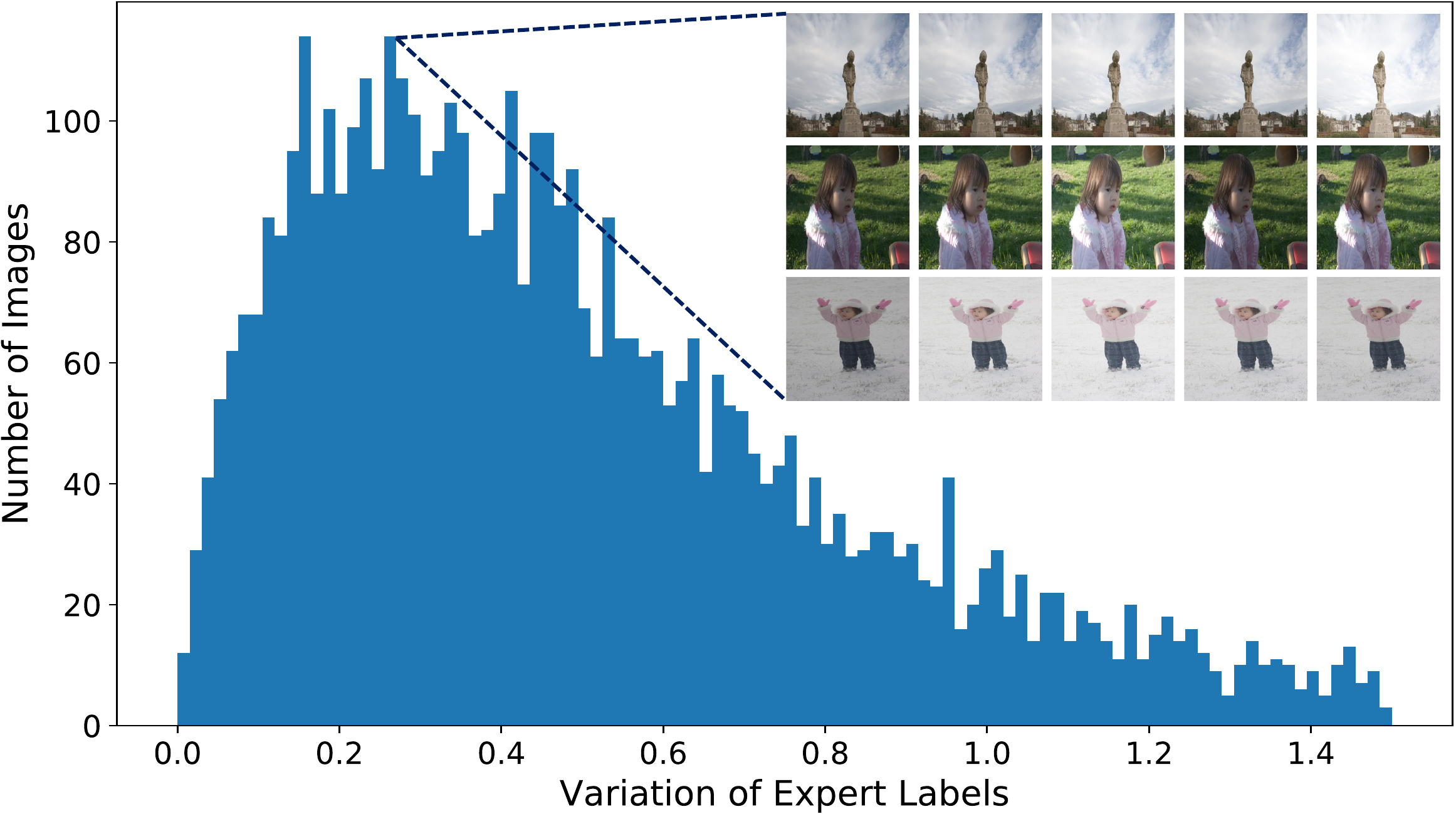}
    \caption{Histogram for the variation of expert labels and some image samples of five experts within the largest bin.}
    \label{fig:fivek_expert_var}
\end{figure}

The MIT FiveK dataset contains 5000 RAW images, each of which was retouched by five experts using Lightroom. In our work, we consider exposure adjustments only. Figure~\ref{fig:fivek_expert_var} shows a histogram that illustrates the variation of expert labels for each image. We count only images with variation below $1.5$, which cover $95\%$ of the whole dataset. We also sample images for five experts within the largest bin for the variation of $0.255$ in our settings; they exhibit significant differences in exposure adjustment. In the dataset, about $75\%$ of the images have variation larger than $0.255$. This analysis lends credence to our claim that exposure personalization is important under the MIT FiveK dataset.

Given the RAW format, it is straightforward to synthesize 20 new images by varying $\Delta_{EV}$ for each original image in Lightroom. However, we use the final processed images as the training data, because ultimately we want the final processed (rather than RAW) image to be personalized.

We randomly sample 500 of original images for testing; for each image, since we know the ground-truth exposure adjustment for each of the five experts, validation is fairly straightforward. We use the other 4,500 images for training. However, unlike supervised learning which directly uses the ground-truth as the signal to train the network, we instead use \textit{indirect} feedback signals in the form of exposure change $\Delta_{EV}$.

$\Delta_{EV}$ is computed as the difference in exposure between the current (synthetic) exposure and that of ground-truth. The synthetic image will be annotated as well-exposed if its $\Delta_{EV}$ is very close to the ground-truth (when the difference is less than 0.1). Otherwise, it will be labeled as either over-exposed or under-exposed based on the deviation. The mapping of reward values is given in Figure~\ref{fig:rlrwd}.
We train a separate model for each expert, and then report their respective performance on the test set.


\subsection{iPhone 7 and Google Pixel Datasets for Native Camera Enhancement}

To further evaluate the effectiveness of our approach, we conducted offline simulations on two popular smartphones, namely iPhone 7 and Google Pixel. The goal of this experiment is to demonstrate that our system can be used to enhance the exposure quality for native cameras beyond personalization. The motivation behind this is that for most native cameras, the exposure metering and adjustment are not semantic-aware beyond faces. It is highly desirable to be able to meter the exposure based on generic semantic content.

We captured 26,679 images using an iPhone 7 and 7,056 images using a Google Pixel for a variety of scene content. We randomly sample $90\%$ of images from each set for training and use the rest for validation to avoid over-fitting. As for training, we hired 5 judges to annotate the exposure quality as being under-exposed, well-exposed, or over-exposed; we then consolidated the annotation results through simple majority voting. We instructed the judges to pay more attention to foreground objects they deem important (such as building, pets, face, humans, cars, and plants) when judging the exposure quality\footnote{Although the final annotation is consolidated by voting and not personalized, it is conceptually possible to train based on individual judges as a means for personalizing the exposure. In our work, this was not done because this experiment is to verify if our approach can be used to enhance the native camera for general users.}.


To augment the training set, we again use Lightroom to synthesize 9 more images with $\Delta_{EV}$ from $-1.0$ to $1.0$ with a step size $0.25$ for each original image. We then fine-tune the basic model using all these images as well as the feedback signals through the reinforcement learning module. The fine-tuning is done separately on the iPhone 7 and Google Pixel datasets to generate two different models. We then deployed them to the respective smartphones and performed comparisons with native camera performance through a user study.




\section{Implementation Details}

Before we report our results, we provide more details on system implementation, namely on network topology, training, and evaluation.

\subsection{Network Topology}

The network topology of our system consists of the backbone network and \changed{adaptive metering} module. We use a trimmed SqueezeNet~\cite{SqueezeNet} as the backbone network to achieve a good trade-off between accuracy and run-time speed. For SqueezeNet, we keep the layers from the bottom up to ``fire7'' and discard the other layers. We then add a dropout layer with dropout ratio as $0.5$ to reduce the risk of over-fitting.

Our \changed{adaptive metering} module is designed to predict soft metering regions so as to generalize the default hardware metering mode. Sitting on the top of backbone net, our \changed{adaptive metering} module branches out into two small subnets; one is used to predict the \changed{importance} map while the other predicts the local exposure adjustment map. We use a 1x1 convolution layer followed by \textbf{TanH} and \textbf{Sigmoid} activation functions, respectively (Figure \ref{fig:overview}). These two maps are then element-wise multiplied before being applied to the global average pooling layer for the final result.


\subsection{Training, Run-time Details and Evaluation Metric}

We use PyTorch to train our system. The input images are all resized to $128 \times 128$, with a batch size of $128$. The labels are normalized within $[-1.0, 1.0]$. All the networks were trained for $35$ epochs with a stepped learning rate (denoted as $r$) policy; $r$ is reduced by half every $15$ epochs. For supervised pre-training, we use the SGD solver with the momentum of $0.9$, weight decay of $0.0002$, and learning rate of $0.003$. For reinforcement learning, we use the Adam solver with an initial learning rate of $0.0001$.

During run-time, once we get the new $EV$ to apply, we decompose $EV$ into ISO($s$) and Exposure duration time($t$) according to the equation $EV=\log_2{t*s}$, as defined in Appendix. To avoid noises, we always prefer a small ISO whenever possible. However, when the duration time exceeds the hardware limit, we will keep increasing ISO to the next level until we find a duration time that supported by the hardware. Note that, the duration time is equivalent to another commonly used concept "Shutter Speed" in here.

Since we are dealing with a regression task, we choose the Mean Absolutely Error (\textbf{MAE}) as the evaluation metric:
\begin{equation}\label{equ:metric}
MAE=\sum_{i=1}^{N}|\Delta_{EV_{i}}-Y_{i}|,
\end{equation}
where $N$ is the number of testing images, and $EV_{i}$ and $Y_{i}$ are the prediction and ground-truth labels of image $i$, respectively.

\section{Experimental Results}
\label{sec:experiment}


\changed{In this section, we first show how our system performed on exposure personalization using the MIT FiveK dataset. Results on this dataset demonstrate that our system is able to learn exposure personalization from five experts individually. For real-time deployment, it is ideal to conduct an extensive deployment of on-line learning for personalization on devices. However, it requires a significant amount of engineering effort to develop such a system as well as a sufficiently large number of user feedback signals for empirical validation. This type of evaluation is hard to be conclusive due to the lack of ground truth. By comparison, the MIT FiveK dataset allows for a more controlled validation process via known ground truth. 
In Section~\ref{sec:exp_enhance}, we evaluate the ability of our system to enhance native camera exposure on two popular smartphones through a user study.}

\subsection{Results on MIT FiveK}
\label{sec:exp_fivek}

As mentioned earlier, the MIT FiveK dataset is used as proof-of-concept for personalization-based reinforcement learning. It is also used for evaluating different parts of our system.

\begin{table}[h]
\centering
\begin{tabular}{ccccccc}
\toprule
   & Mean & ExpA & ExpB & ExpC & ExpD & ExpE \\
\midrule
 RL & 0.2457 & 0.3160 & 0.2917 & 0.3526 & 0.3069 & 0.3144 \\
 RL+PT & \textbf{0.2019} & \textbf{0.2936} & \textbf{0.2349} & \textbf{0.2725} & \textbf{0.2491} & \textbf{0.2518} \\
\bottomrule
\end{tabular}
\caption{Testing MAE comparison between reinforcement learning from ImageNet (RL) and reinforcement learning with our pre-training (RL+PT).}
\label{tab:fivek_finetune_comp}
\end{table}

\subsubsection{Evaluation of Supervised Pre-training}

Without supervised pre-training, one can choose to fine-tune the whole network by pre-training based on the classification task using ImageNet. However, since the \changed{adaptive metering} module is not used in any previous classification network, we can only borrow the pre-trained weights in the backbone network from those pre-trained on ImageNet (such as the pre-trained SqueezeNet used in our experiments). All the other new layers are initialized with standard zero mean and 0.01 standard deviation Gaussian function.

Table~\ref{tab:fivek_finetune_comp} shows the effect of fine-tuning from ImageNet and fine-tuning from our supervised pre-trained weights for different personalized models. The MAE numbers support our design decision to use supervised pre-training, since they are reduced for all five personalized models. This is likely because the pre-training step allows the model to learn more relevant representations tailored for exposure control, and not generic features used for image classification.

\subsubsection{Evaluation of \changed{Adaptive Metering} Module}

\begin{figure}[h!t]
    \centering
    \includegraphics[width=\linewidth]{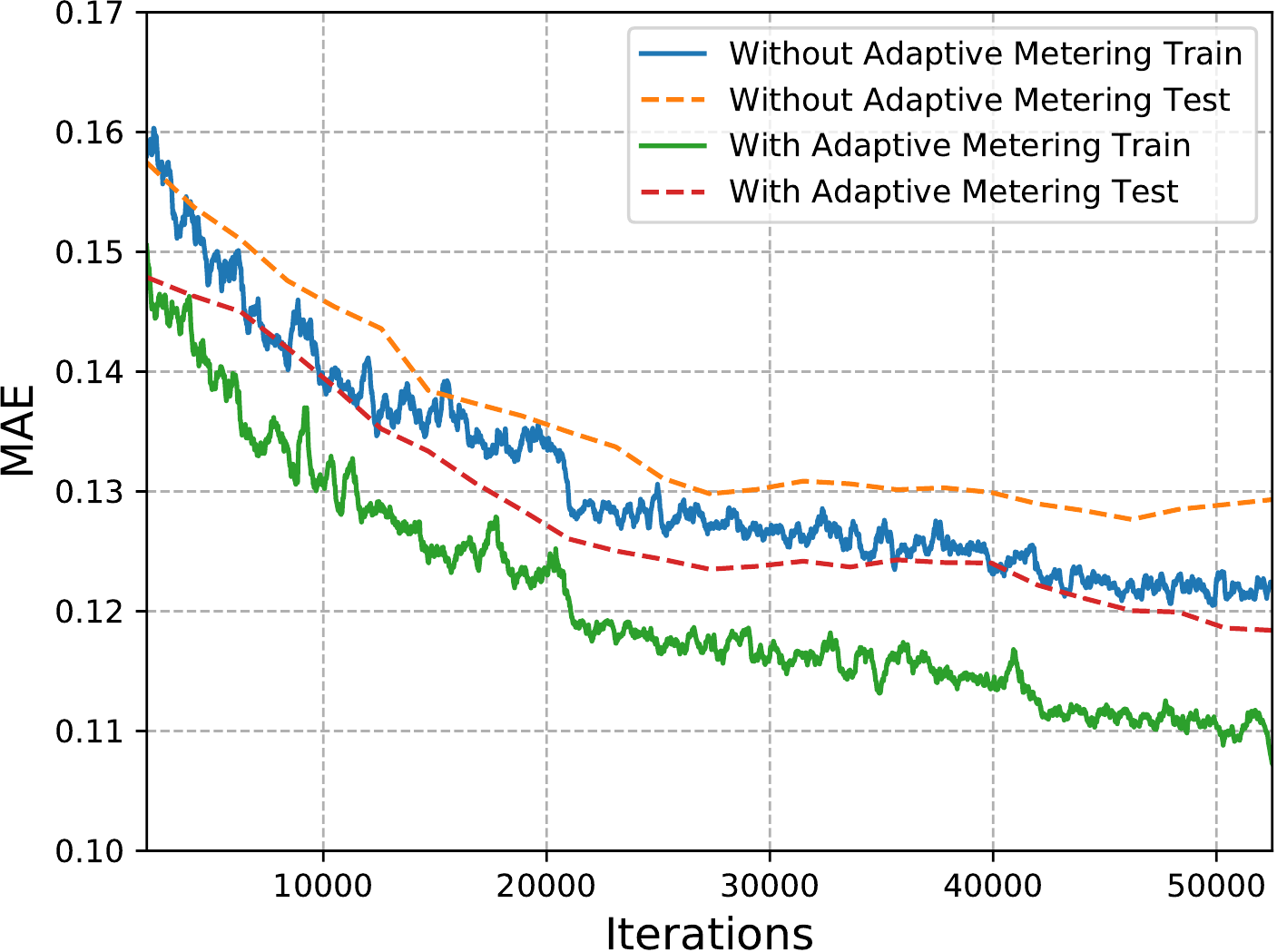}
    \caption{Training and testing performance graphs that show the effect of using the \changed{adaptive metering} module.}
    \label{fig:base_vs_attn_curve}
\end{figure}

The goal of our proposed \changed{adaptive metering} module is to generalize the heuristic hardware metering options by adaptively predicting the \changed{importance} map for each image. The baseline approach is simply outputting the exposure map only, treating each local region with same weight for aggregating the final exposure adjustment. As comparison, we remove the \changed{importance} branch and the subsequent element-wise multiple layer and trained a baseline model. Figure~\ref{fig:base_vs_attn_curve} shows that during the pre-training stage, adding the \changed{adaptive metering} module significantly reduces the regression error for both the training and testing. This appears to indicate that the \changed{importance} branch learns some meaningful maps for prioritization of certain local areas at the expense of others.


To further illustrate the effectiveness of the \changed{adaptive metering} module, we show both exposure map and \changed{importance} map for the testing images under different scene content and lighting. For visualization, the exposure map is shown in pseudo-color while the intensities of the \changed{importance} map is linearly converted to $[0, 255]$. The red regions in the exposure map represent over-exposed areas (where exposure needs to be reduced) while the green regions represent under-exposed areas (where exposure needs to be increased). As for the \changed{importance} map, the brighter the region is, the higher the priority for exposure adjustment.

\begin{figure*}[!t]
    \centering
    \includegraphics[width=\linewidth]{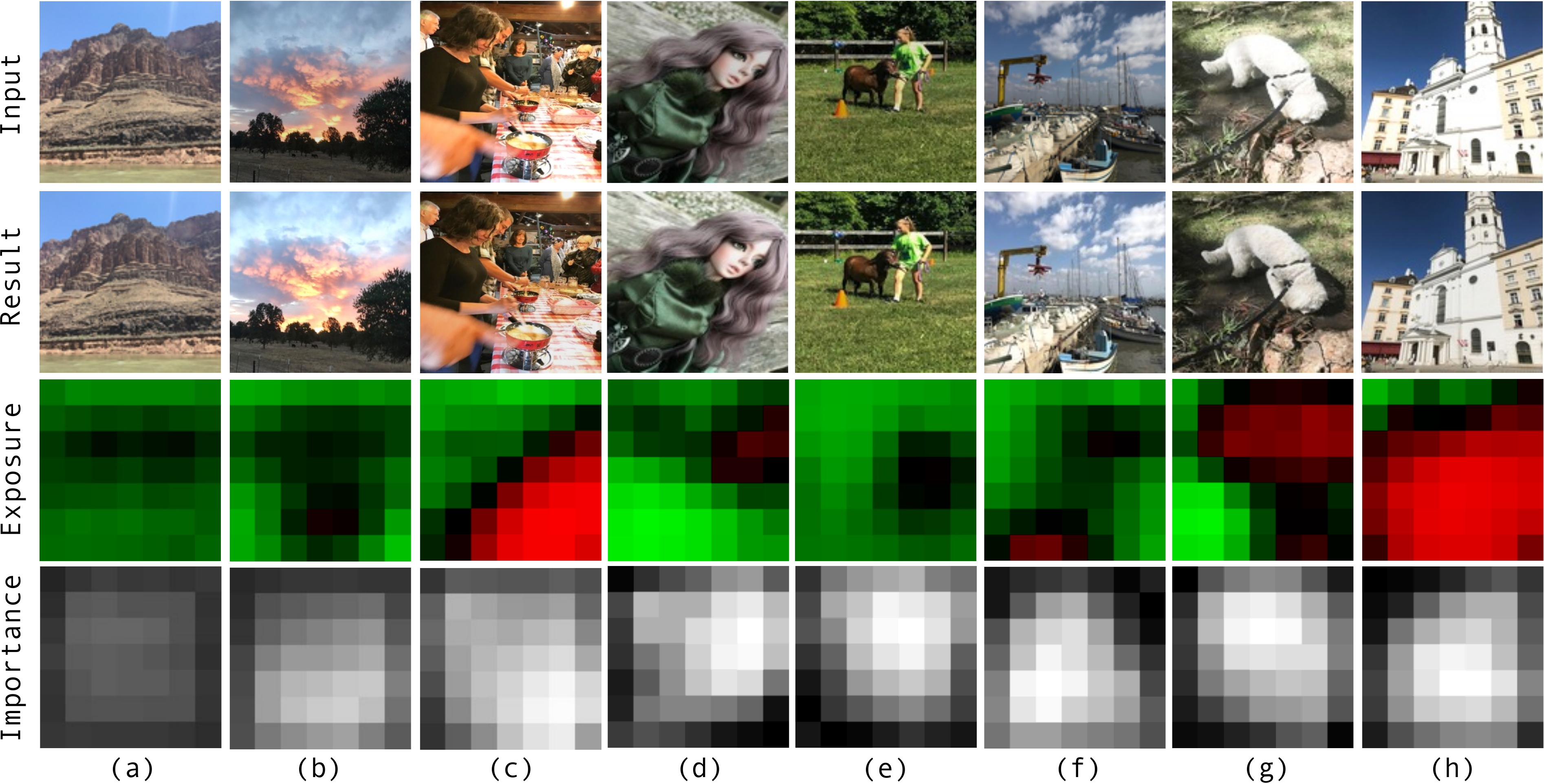}
    \caption{Exposure and \changed{importance} maps on representative images. For image (a), the foreground occupies most of the scene; here, the \changed{importance} map is close to being uniform. For image (b-c), there are no objects with obvious semantics, for such cases, the \changed{importance} maps are focusing on the region exposed incorrectly. For image(d-h), objects with obvious semantics can be found in the scene which the \changed{importance} map should focus on such regions accordingly.}
    \label{fig:attn_vis_general}
\end{figure*}

Figure~\ref{fig:attn_vis_general} shows the exposure and \changed{importance} maps for representative images. Our method appears to be able to predict semantically reasonable exposures as shown in the second row. Notice that for image (a), where there is no discernible local foreground of interest, the maps are close to being uniform. For the other images with local foregrounds of interest, the maps are substantially more unevenly distributed. Visually, they correlate with objects of interest. In particular, for image (f), our system deems the building to be over-exposed; there are more details on the facade after exposure adjustment.

\begin{figure}[!t]
    \centering
    \includegraphics[width=\linewidth]{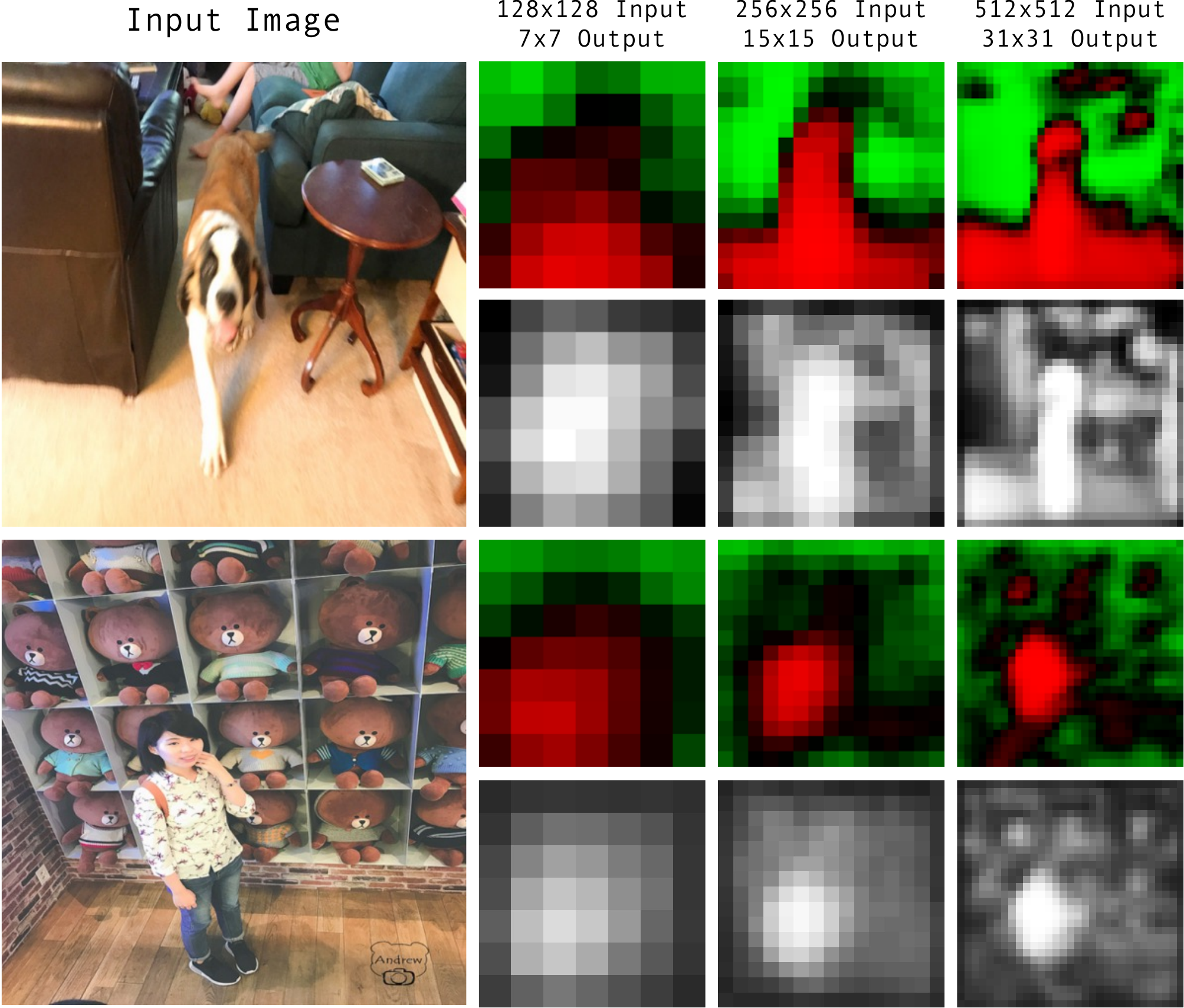}
    \caption{Exposure and \changed{importance} maps for three different input sizes ($128 \times 128$, $256 \times 256$, and $512 \times 512$). The corresponding output feature map sizes are $7 \times 7$, $15 \times 15$, and $31 \times 31$.}
    \label{fig:map_diff_size}
\end{figure}

Our network is based on FCN (fully convolutional networks), which can be easily modified to accommodate larger input and output feature maps that are capable of storing more details. Figure~\ref{fig:map_diff_size} shows the exposure and \changed{importance} maps associated with input sizes of $128 \times 128$, $256 \times 256$, and $512 \times 512$. Notice the increasingly better detail with higher input size; this is evidence that our system is able to learn scene semantics.

Interestingly, even without direct supervision, our end-to-end system is capable of learning the latent \changed{importance} maps. This may be attributed to our local-to-global aggregation design through both the element-wise multiple layer and global average pooling layer. This suggests that an adaptive weighted \changed{importance} map is effective in generalizing the heuristic hardware metering modes.

\subsubsection{Evaluation of Reinforcement Learning}

Table~\ref{tab:fivek_reinforce_comp} compares the performance of fully supervised learning and reinforcement learning using the MIT FiveK database for training and testing. Fully supervised learning is done by directly regressing $\Delta_{EV}$ using ground-truth data, and its numbers can be considered as upper-bound performance. The numbers for reinforcement learning are very competitive even though it makes use of coarse-grained feedback signals (``over-exposed'',``under-exposed,'' or ``well-exposed'').

In addition to the five personalized models available from the MIT FiveK database, we also compute the average $\Delta_{EV}$ of all the five ground-truth labels for each image. We train a ``mean'' model that to represent the preference of the average person. We then use these six trained models to run the test set for each expert (same testing images but different $\Delta_{EV}$ labels). Unsurprisingly, results shown in Table~\ref{tab:fivek_personal_comp} indicate that the personalized model trained for each expert generates the smallest error for that same expert. This demonstrates the necessity for personalization of exposure adjustment.




\begin{table}[h]
\centering
\begin{tabular}{ccccccc}
\toprule
   & Mean & ExpA & ExpB & ExpC & ExpD & ExpE \\
\midrule
 FSL & 0.1926 & 0.2770 & 0.2272 & 0.2688 & 0.2412 & 0.2450 \\
 RL & 0.2019 & 0.2936 & 0.2349 & 0.2725 & 0.2491 & 0.2518 \\
\bottomrule
\end{tabular}
\caption{Testing MAE for fully supervised learning (FSL) and reinforcement learning (RL).}
\label{tab:fivek_reinforce_comp}
\end{table}

\begin{table}[h]
\centering
\begin{tabular}{ccccccc}
\toprule
& & \multicolumn{5}{c}{Test}\\
& & ExpA & ExpB & ExpC & ExpD & ExpE \\
\midrule
\multirow{6}{*}{\begin{sideways} Train \end{sideways}} & Mean & 0.3862 & 0.2483 & 0.3355 & 0.2702 & 0.2919 \\
& ExpA & \textbf{0.2936} & 0.3659 & 0.4897 & 0.3952 & 0.4433 \\
& ExpB & 0.4143 & \textbf{0.2349} & 0.3037 & 0.2588 & 0.2731 \\
& ExpC & 0.5057 & 0.2816 & \textbf{0.2725} & 0.2703 & 0.2706 \\
& ExpD & 0.4008 & 0.2506 & 0.3151 & \textbf{0.2491} & 0.2791 \\
& ExpE & 0.4679 & 0.2588 & 0.2881 & 0.2645 & \textbf{0.2518} \\
\bottomrule
\end{tabular}
\caption{Testing MAE matrix showing the effect of applying a trained model on testing data with different ground-truths (depending on the expert).}
\label{tab:fivek_personal_comp}
\end{table}

To further show that our model is able to learn personalized exposure preferences, we run our model for each expert on the test set from the MIT FiveK dataset. We compute the percentage of images which $\Delta_{EV}$ predicted by our model is nearest to the ground-truth $\Delta_{EV}$ (compared to the ground truth for the other four experts); the results are shown in Table~\ref{tab:fivek_personal_comp2}. There are two interesting observations. First, for any row or column, the highest percentage is where the prediction and ground truth are from the same expert. Second, there is significant variation in percentages across the table. These observations support the existence of exposure preference in the database and show that our method is able to reasonably capture these preferences.

\begin{table}[h]
\centering
\begin{tabular}{ccccccc}
\toprule
& & \multicolumn{5}{c}{Percentage of Images}\\
& & ExpA & ExpB & ExpC & ExpD & ExpE \\
\midrule
\multirow{5}{*}{\begin{sideways} Train \end{sideways}}
& ExpA & \textbf{67\%} & 13\% & 2\% & 13\% & 4\% \\
& ExpB & 10\% & \textbf{42\%} & 13\% & 19\% & 16\% \\
& ExpC & 3\% & 16\% & \textbf{52\%} & 13\% & 16\% \\
& ExpD & 14\% & 18\% & 14\% & \textbf{38\%} & 15\% \\
& ExpE & 3\% & 19\% & 15\% & 14\% & \textbf{48\%} \\
\bottomrule
\end{tabular}
\caption{Accuracy of prediction from one expert given the ground truth (trained) from another expert. Each number in the table is the percentage of images which the predictions of one model is nearest to the ground-truth $\Delta_{EV}$. For example, only 2\% of the images predicted by ExpC are closest to the ground truth from ExpA.}
\label{tab:fivek_personal_comp2}
\end{table}

\begin{figure*}[t!]
    \centering
    \includegraphics[width=\linewidth]{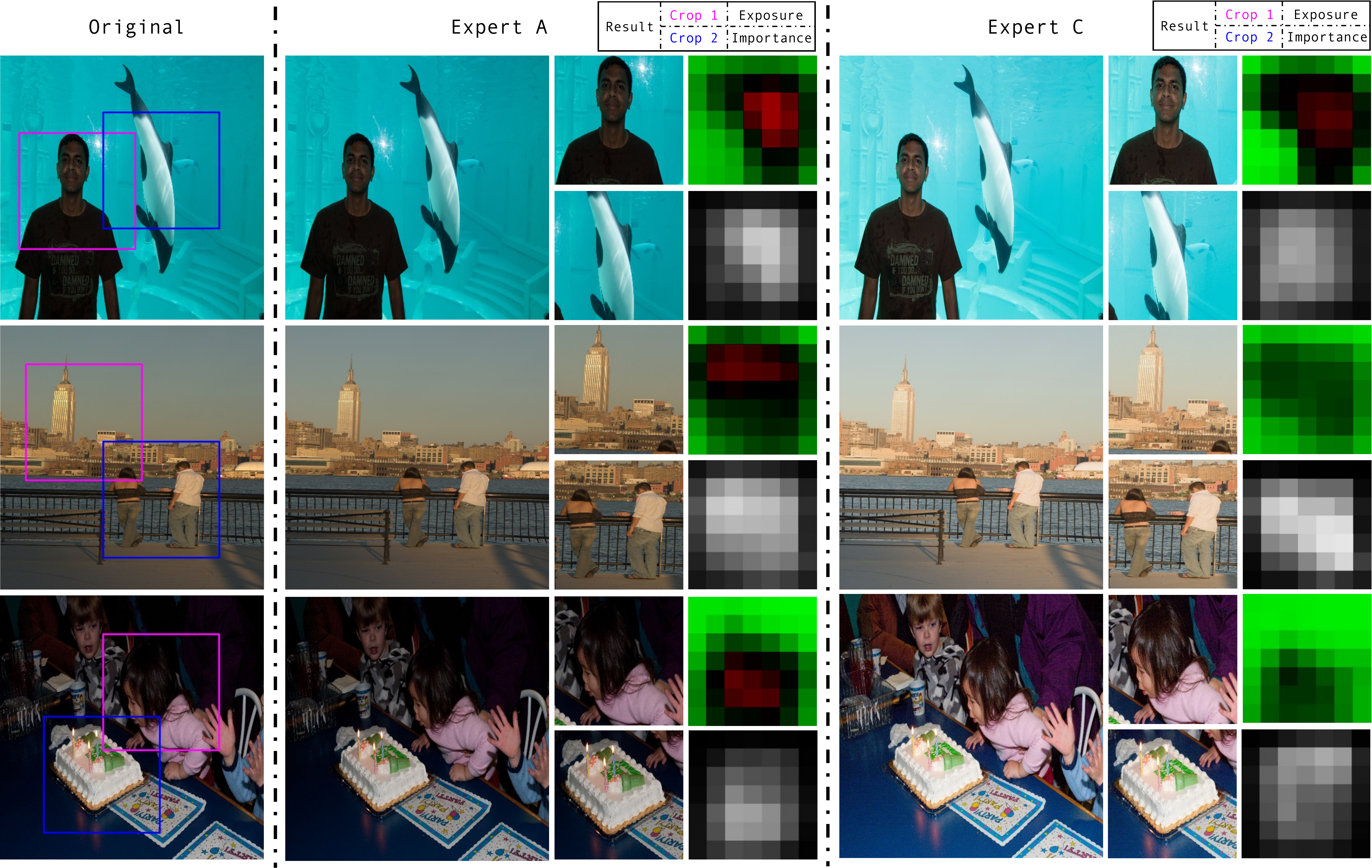}
    \caption{Exposure and \changed{importance} map visualization of testing examples for experts A and C. Note that for each example, the exposure and \changed{importance} maps are for the entire image.}
    \label{fig:fivek_exp_sal_comp}
\end{figure*}

Figure~\ref{fig:fivek_exp_sal_comp} shows results for two experts (A and C) who have significant differences in exposure preferences. For each example, we show close-ups of two regions to highlight the difference in exposure preference in the third and sixth columns. The different exposure preferences are reflected in the different exposure and \changed{importance} maps shown in the fourth and seventh columns.


\begin{figure}[!b]
    \centering
    \includegraphics[width=\linewidth]{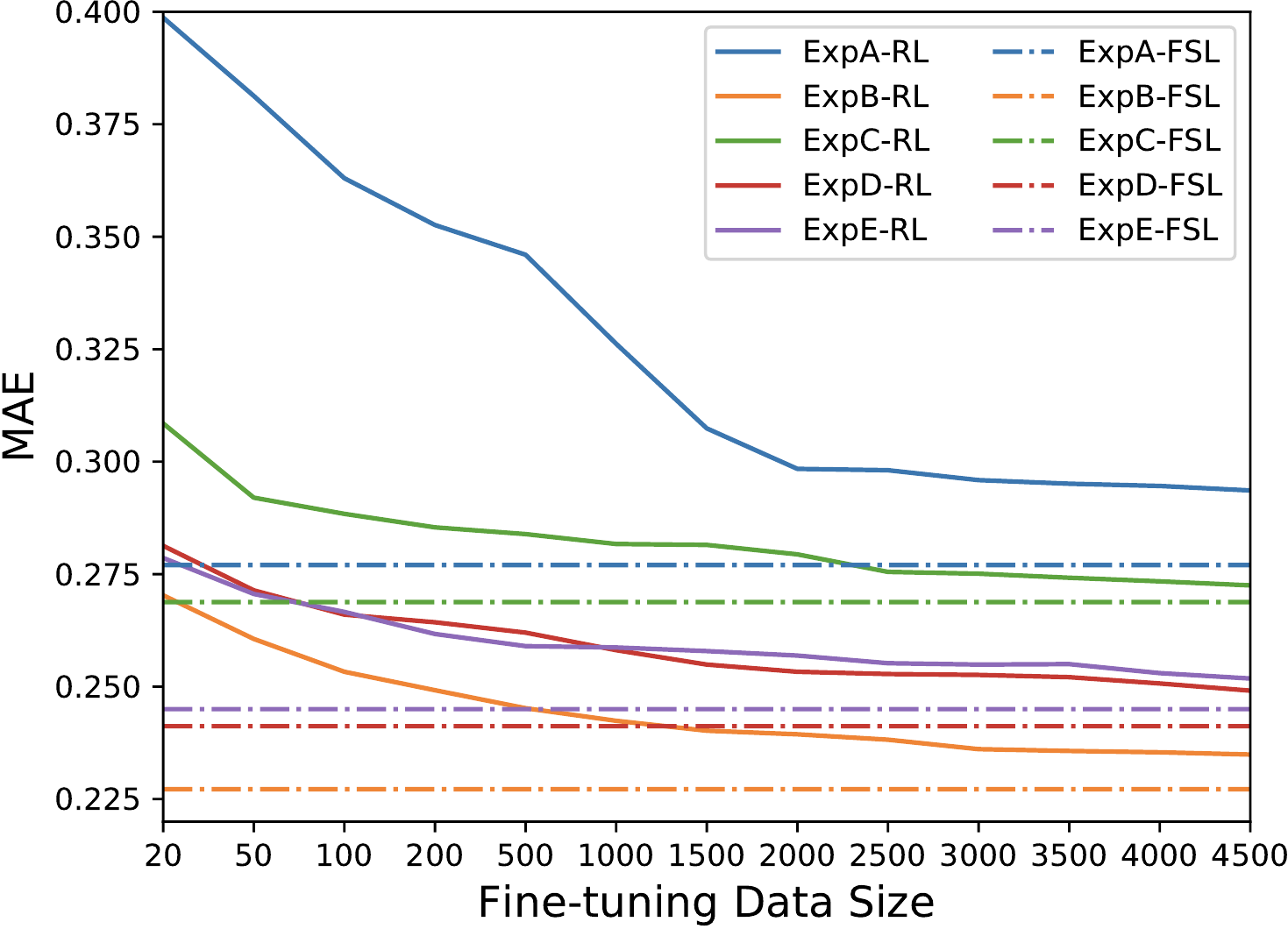}
    \caption{Testing MAE as a function of training data size. The solid lines represent performance of our reinforcement learning based method while the dotted lines are the reference (lower-bound) performance based on fully supervised learning.}
    \label{fig:fivek_datasize_comp}
\end{figure}

We also evaluate the performance of reinforcement learning with respect to the size of training data. Figure~\ref{fig:fivek_datasize_comp} shows that the regression error monotonically decreases with increase of training data size, which is expected. Each solid line represents the testing performance of each personalized model trained by reinforcement learning, while each dotted line denotes the corresponding testing performance of the reference model that was trained by fully supervised learning with all 4,500 training images. Each dotted horizontal line represents upper bound performance. The errors seem to drop more quickly when the images are fewer than 200 for experts A, B, C, D, and E. Also, after increasing the number of images to a few hundred, the performance of our RL-based method approximates that of fully supervised learning.


\subsubsection{Comparison with Other Methods}


Table~\ref{tab:fivek_other_cmp} compares our performance in terms of MSE (mean squared error) against those for the systems of Hwang et al. 2012~\cite{hwang2012context} and Yan et al. 2016~\cite{Yan:2016} (both of which are post-processing techniques with spatially-varying operations, and both report MSE as quantitative objective evaluation). In particular, Yan et al.~\cite{Yan:2016} use a set of hand-crafted and deep learning features to learn spatially-variant local mapping functions and predict the final L channel value of each pixel in CIE*LAB space. In contrast, our method is more light-weight and can be considered as a global approach. It is noted that MSE is not a good metric for perceptual quality \cite{hwang2012context,Exposure:hu}. Regardless, our results are still competitive, with the advantages of our system being geared for personalization and allowing real-time capture. In addition, our system is trained without the need of full supervision, which is more scalable for practical deployment.

\begin{table}[t!]
\centering
\begin{tabular}{ccc}
\toprule
Method & Ran.250 (L,a,b) & H.50 (L,a,b)\\
\midrule
Hwang et al. 2012~\cite{hwang2012context} & $15.01$ & $12.03$\\
Ours (RL) & $13.51$ & $11.21$\\
Yan et al. 2016~\cite{Yan:2016} & $9.85$ & $8.36$\\
\bottomrule
\end{tabular}
\caption{Comparison of MSE errors obtained with our method (RL) and previous methods (fully supervised learning) on the MIT-Adobe FiveK database.}
\label{tab:fivek_other_cmp}
\end{table}

\subsection{Results for Exposure Control Enhancement}
\label{sec:exp_enhance}



As proof-of-concept for improving exposure control though semantic-awareness (beyond faces), we trained two different models, one for iPhone 7 and the other for Google Pixel. The training was done using their own respective datasets.

We then developed a test app that is deployed to both iPhone 7 and Google Pixel. We can control the exposure only through API calls and not directly through the firmware. Not surprisingly, we noticed there is latency in the API calls; the amount of latency is 3-5 frames. Our results show that even though the models were trained using synthetic datasets, our exposure control system is still able to achieve good trade-off among the steadily state, temporal oscillation, and run-time speed. On iPhone 7, our model takes 17 ms to predict the exposure; on Google Pixel, it takes around 27 ms.

For performance comparison, we mounted two phones side-by-side on rig and captured pictures simultaneously. To address the view difference issue, we followed the approach proposed by \cite{Ignatov2017DSLR,Wang:2011:EIC}) to first align two images by local features and then crop for best comparison.

\begin{figure*}[t!h]
    \centering
    \includegraphics[width=\linewidth]{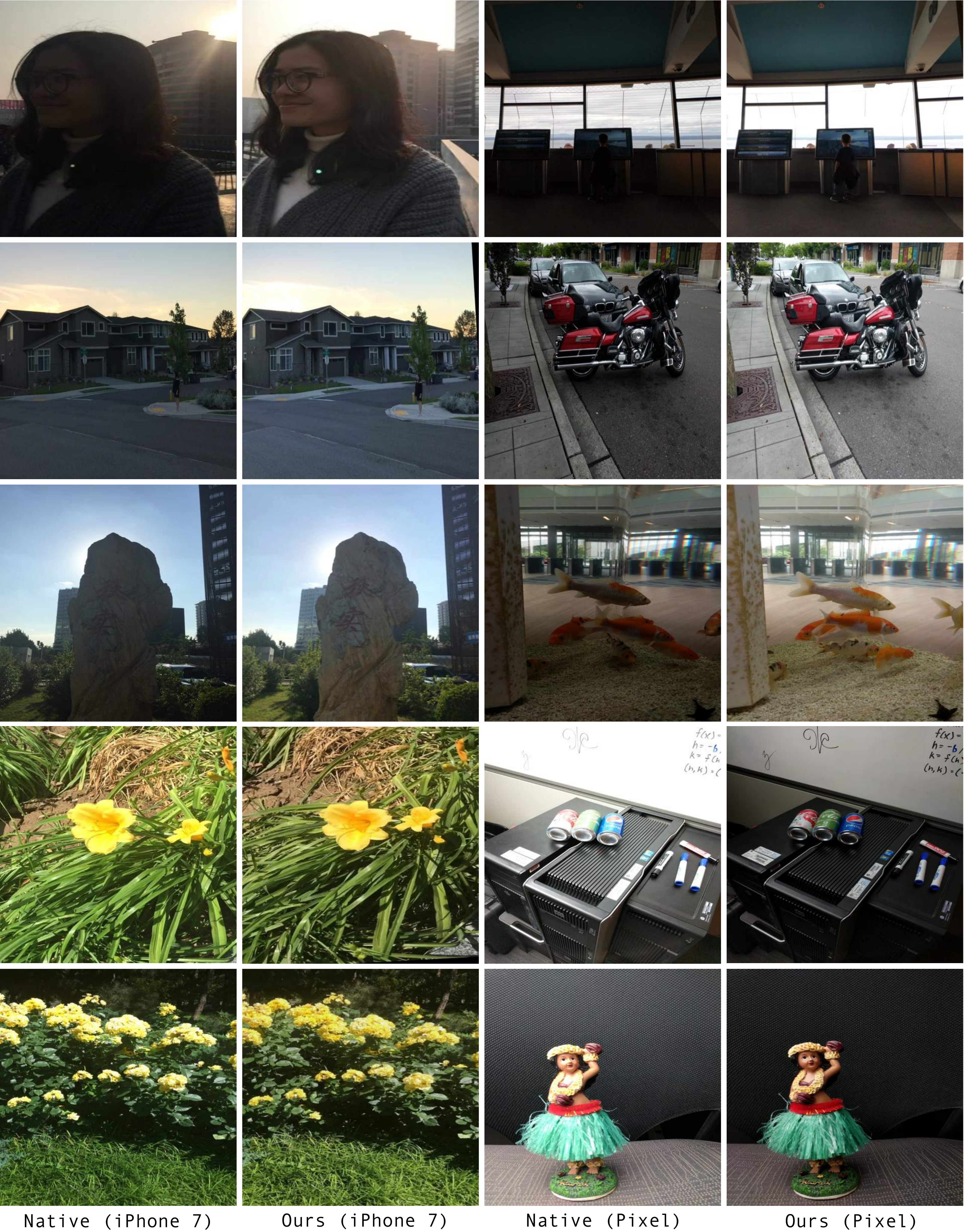}
    \caption{Side-by-side capture of various scenes using the native camera apps of iPhone 7 and Google Pixel and ours. Our system tend to better expose foreground objects.}
    \label{fig:enhance_pairwise}
\end{figure*}

Figure~\ref{fig:enhance_pairwise} shows a few comparisons with iPhone 7 and Google Pixel. When the face is not frontal, iPhone 7 typically fails to detect it, leaving it under-exposed in a backlit scene. Our model is able to better expose foreground subjects at the expense of slightly over-exposing the background. A similar assessment can be made for Google Pixel (e.g., the scene with the person with the back facing the camera). The native camera apps for both iPhone 7 and Google Pixel tend to optimize over the whole image, and as a result, under-expose the foreground for scenes with bright backgrounds. Our system is able to prioritize exposure for non-person objects as well, e.g., the building and motorcycle in the second and third rows. For the examples in the last two rows, our system instead reduces the exposure for better detail. This is in comparison to the over-exposed versions from iPhone 7 and Google Pixel's native camera app.
These examples illustrate the ability of our model to learn scene semantic representations for better exposure control.


\begin{figure*}[!ht]
    \centering
    \includegraphics[width=0.9\linewidth]{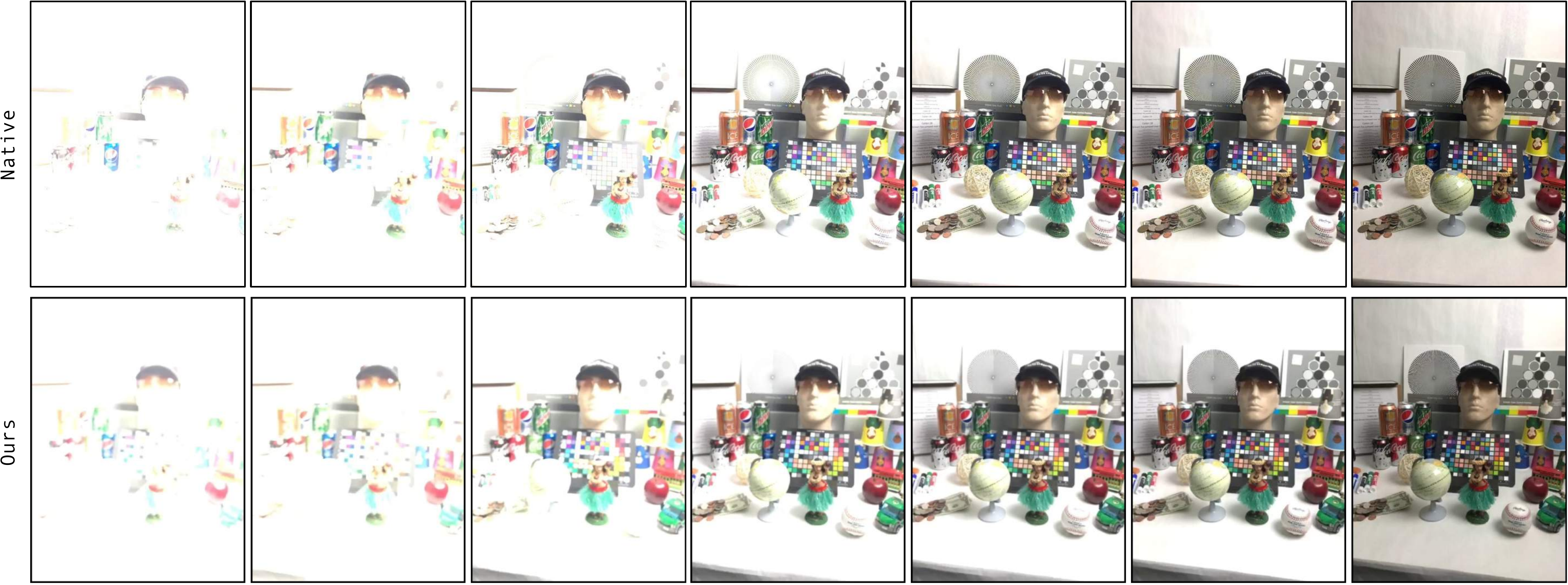}
    \caption{Comparison of exposure dynamics between our model and the iPhone 7 native camera (both in response to a sudden increase in lighting).}
    \label{fig:exposure_dynamics}
\end{figure*}

Figure~\ref{fig:exposure_dynamics} shows snapshots that represent the response of our system to abrupt lighting changes. The dynamics of exposure varying mimics the native camera app well. Please refer to supplemental videos for more examples of how our deployed real-time exposure control system responds in  real-time to camera motion in different scenes and changing lighting conditions.

\begin{figure}[!h]
    \centering
    \includegraphics[width=\linewidth]{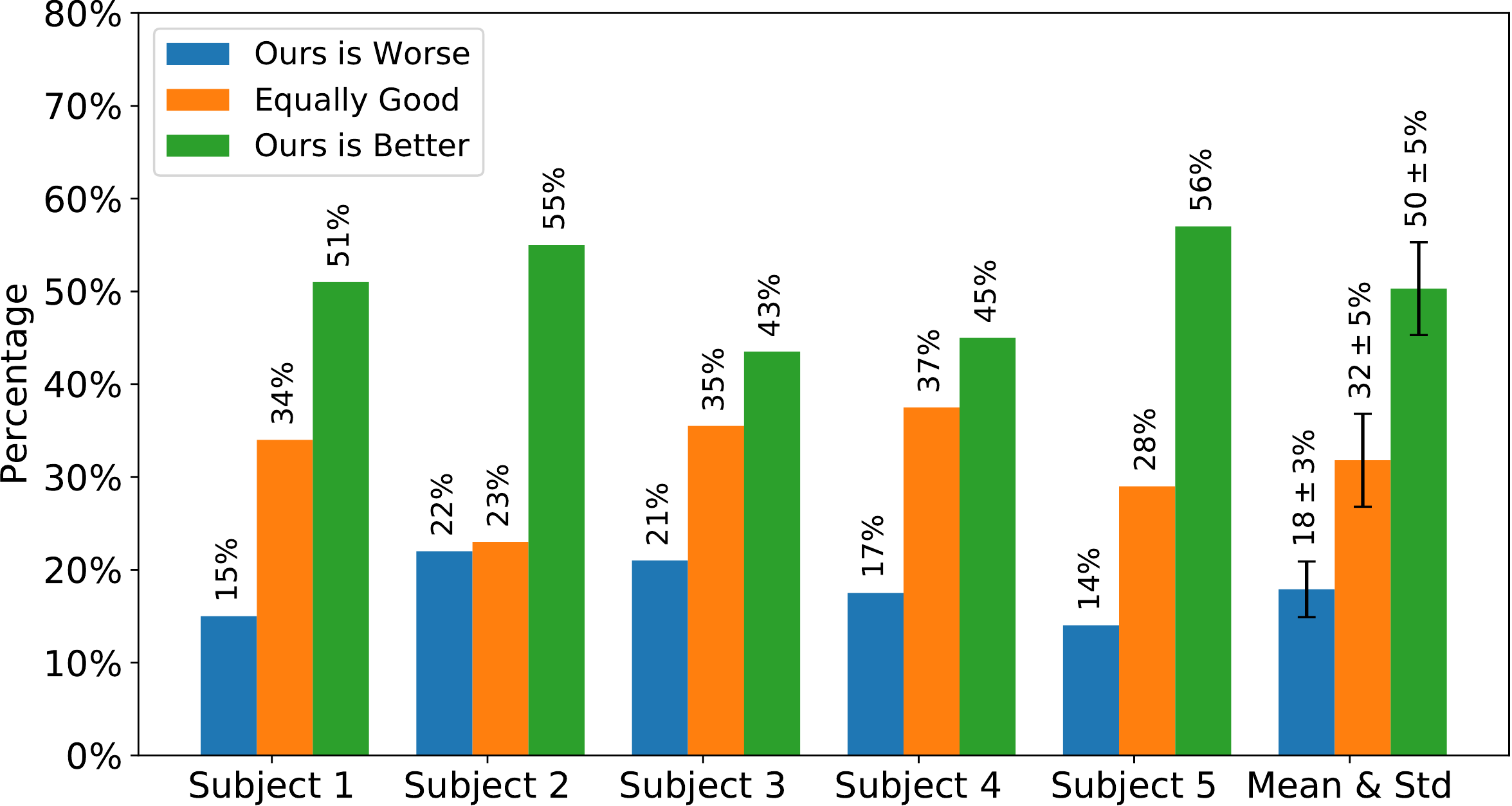}
    \caption{Histogram showing preferences for each of the five subjects as well as the mean result with standard deviation.}
    \label{fig:user_study}
\end{figure}
\subsubsection{User Study}

We performed a user study to compare performance of our system with that of the native camera app. To do this, we first captured a 100 pair of images using two iPhone 7s placed side-by-side on many different scenes; one is running the native camera while the other is running our app. Similarly, we   captured another 100 pairs of images for side-by-side comparison on two Google Pixel phones. In the user study, a subject is shown a pair of images, one being our result and the other from the native camera app. The images are randomly arranged. In this A-B test, the subject is asked to select amongst the choices ``Left is better", ``Right is better," or ``Equally good.''

There are five subjects in the study, with each looking at the same 200 image pairs. The results, shown in Figure~\ref{fig:user_study}, validate our design decisions for a reinforcement learning based exposure control system. These results are also significant in that the subjects in the user study were not involved in annotating the training data.

To further analyze the variation of five subjects for each image pair, we evaluate the consistency of five subjects for each image pair in Table~\ref{tab:user_study_var}. In this table, $N_{consistency}$ is used to denote the maximum number of subjects (up to five) having the same label for each image pair. For example, $N_{consistency} = 4$ means four subjects having the same opinion for an image pair. $81\%$ of the image pairs have at least four subjects agreeing with each other.

\begin{table}[h]
\centering
\begin{tabular}{ccccccc}
\toprule
$N_{consistency}$ & Percentage \\
\midrule
5 & 55\% (110/200)\\
4 & 26\% (52/200)\\
3 & 18\% (36/200)\\
$\le 2$ & 1\% (2/200)\\
\bottomrule
\end{tabular}
\caption{The percentage of image pairs with different $N_{consistency}$.}
\label{tab:user_study_var}
\end{table}

\changed{We also evaluate the variation of five subjects through Kendall's coefficient of concordance\cite{KendallW} which denotes as $W$ here. In order to show the result of all 200 image pairs, we take an average of $W_i$ that computes from each image pair respectively. The final averaged $W=0.67$ which shows strongly concordancy among five subjects.}





\section{Discussion}

In this section, we discuss a number of issues on system design and feature extensions.

\subsection{Training on HDR or DSLR Images}
In principle, our proposed system can be trained on HDR images, in which case, $f$ has to take a RAW image as input, with the corresponding exposure quality label judged based its the processed image. However, in practice, accessing HDR and RAW images is challenging on most mobile devices, so collecting the RAW (HDR) images is not as easy as collecting JPEG images for training data. At runtime, the same type of image would need to be accessed in real-time; even if the RAW image is accessible, it would be done at the expense of performing the ISP (image signal processing) pipeline in software, which would be unacceptably slow.

For the DSLR images, since each device's ISP pipeline is different, for the same RAW image and equivalent exposure settings, the output is very likely to be different. As a result, learning from RAW data from one device and applying it to another (in our case mobile cameras) would be problematic.

\subsection{Dealing with HDR Content}
The typical scene is inherently of high dynamic range, and it is better to capture HDR if possible. This is especially true for backlit or frontlit scenes. It is certainly possible to incorporate some of our ideas to select the bracketing exposures for optimal semantic-aware HDR capture. Given that sequential capture is required for HDR, the problem is significantly more challenging due to possible camera and/or scene motion. HDR also requires post-processing, which is outside the scope of our work.

\subsection{Camera 3A Integration}
In this work, we wish to demonstrate the most obvious feature of the phone in live viewfinder mode, namely exposure. In practice, auto-exposure works in conjunction with auto-focus and auto-white balance, collectively termed the \textit{camera 3A}. The amount of training data that are required to account for 3A would be substantially larger, but the principle would be similar. An end-to-end deep learning based 3A system (given the recent advances in color constancy~\cite{FC4:2017}) is future work.

\subsection{Incorporate with User Input}
Our system is designed to predict the optimal exposure to obviate the need for user manual interaction, such as ``Tap-and-Tweak''. It is possible to add manual overrides through manual area specification through lassoing or tapping on an image area. In principle, our system should still work by replacing the learned \changed{importance} map with a heuristically defined \changed{importance} map based on user selection.

%% file: conclusion.tex
\section{Conclusions}

In this paper, we propose a new semantic-aware exposure control system based on reinforcement learning. Our system consists of three major components: (1) supervised pre-training to mimic the native camera's control behavior, (2) an \changed{adaptive} metering model, and (3) a reinforcement learning model that can be trained for both personalization and enhancing the native camera capabilities. We conducted extensive experiments on MIT FiveK and our own captured datasets. The encouraging results validate our system design. It would be useful to be able to decide on the need to change exposure time or gain (ISO). In particular, accounting for scene motion is important. As an example, in sports photography, it is common practice to increase shutter speed (i.e., reduce exposure time) while increasing the ISO. Another issue is that of speeding up performance; while this can be accomplished through model compression (e.g., by pruning~\cite{han2015deep_compression} and binarization~\cite{Rastegari2016}), more work is required to ensure that accuracy is not too severely compromised.

%% file: appendix.tex
\section{Camera Exposure} 

According to ~\cite{ISO_2720:1974} and ~\cite{Ray:2000}, the exposure equation is defined as
\begin{equation}\label{equ:exposure}
\frac{LS}{K} = \frac{N^2}{t} ,
\end{equation}
where $N$ is the aperture (f-number), $t$ is the exposure time (``shutter speed") in seconds, $L$ is the average scene luminance, $S$ is the ISO speed, and $K$ is constant (commonly set to $12.5$). For a typical mobile camera, $N$ is fixed as well. 

In photography, $EV$ is a number that represents the combination of $t$ and $N$ such that all combinations that yield the same exposure have the same EV for any fixed scene luminance $L$ and ISO $S$. In practice, $EV$ is defined in base-2 logarithmic scale. Specifically, when applied to the left hand side of Eq.~\ref{equ:exposure}, $EV$ defines the target exposure value and computed as $EV=\log_2{\frac{LS}{K}}$ , while when applied to the right hand side, $EV$ represents the camera setting and is computed as $EV=\log_2{\frac{N^2}{t}}$. The ``correct" exposure is achieved when both versions of $EV$ match. For example, given the same lighting condition, increasing ISO would increase $EV$; this increase in $EV$ can be matched by decreasing the exposure time $t$. Given fixed ISO $S$, decreasing $L$ requires $t$ to be increased. For a typical mobile camera, only ISO and exposure time can be adjusted. By rewriting Eq.~\ref{equ:exposure}, for a fixed scene luminance $L$, we can redefine $EV$ as:
\begin{eqnarray}  
\centering
 EV &= \log_2{\frac{1}{t*S}} + \log_2{(K*N^2)}
\end{eqnarray}

Since the second term $\log_2{(K*N^2)}$ is constant for a camera, for simplicity, we set $EV=\log_2{(t*S)}$. Hence, the exposure adjustment $\Delta_{EV}$ will always involve $t$ and $S$.
